%% file: main.tex
\documentclass{article}

\usepackage{microtype}
\usepackage{graphicx}
\usepackage{subfigure}
\usepackage{booktabs} 

\usepackage{hyperref}

\usepackage[accepted]{style/icml2021}

\input{packages}
\input{macros}

\icmltitlerunning{Scaling properties of deep residual networks}

\begin{document}

\twocolumn[
\icmltitle{Scaling Properties of Deep Residual Networks}



\icmlsetsymbol{equal}{*}

\begin{icmlauthorlist}
\icmlauthor{Alain--Sam Cohen}{id}
\icmlauthor{Rama Cont}{ox}
\icmlauthor{Alain Rossier}{ox,id}
\icmlauthor{Renyuan Xu}{ox}
\end{icmlauthorlist}

\icmlaffiliation{ox}{Mathematical Institute, University of Oxford}
\icmlaffiliation{id}{ InstaDeep}

\icmlcorrespondingauthor{Alain Rossier}{rossier@maths.ox.ac.uk}

\icmlkeywords{Machine Learning, ICML, Deep Learning, Neural networks, Residual Networks, ResNet, Scaling limit, Stochastic Differential Equation, Ordinary Differential Equation, Convolutional network}

\vskip 0.3in
]



\printAffiliationsAndNotice{}  

\begin{abstract}
Residual networks (ResNets) have displayed impressive results in pattern recognition and, recently, have garnered considerable theoretical interest due to a perceived link with neural ordinary differential equations (neural ODEs). This link relies on the convergence of network weights to a smooth function as the number of layers increases. We investigate the properties of weights trained by stochastic gradient descent and their scaling with network depth through detailed numerical experiments. We observe the existence of scaling regimes markedly different from those assumed in neural ODE literature. Depending on certain features of the network architecture, such as the smoothness of the activation function, one may obtain an alternative ODE limit, a stochastic differential equation or neither of these. These findings cast doubts on the validity of the neural ODE model as an adequate asymptotic description of deep ResNets and point to an alternative class of differential equations as a better description of the deep network limit.
\end{abstract}

\input{sec/1_introduction}

\input{sec/2_methodology}

\input{sec/3_experiments}

\input{sec/4_asymptotics}

\input{sec/5_conclusion}
\bibliography{ref}
\bibliographystyle{style/icml2021}
\clearpage
\onecolumn
\input{sec/appendix-final}

\end{document}

%% file: packages.tex
\usepackage{amsmath,amsthm}
\usepackage{mathtools}
\usepackage{amssymb}
\usepackage{amsopn}
\usepackage{dsfont}
\usepackage{color}
\usepackage{hyperref}
\usepackage{url}
\usepackage{tikz}
\usepackage{tabu}
\usepackage{enumerate}
\usepackage{placeins}
\usepackage{eso-pic}
\usepackage{caption}
\usepackage{eucal}

%% file: macros.tex
\usepackage{mathtools}
\usepackage{float}
\PassOptionsToPackage{mathcal}{euscript}
\usepackage{multicol}

\newcommand{\abs}[1]{\left\vert#1\right\vert}
\newcommand{\norm}[1]{\left\lVert#1\right\rVert}
\newcommand{\tensornorm}[1]{{\left\vert\kern-0.25ex\left\vert\kern-0.25ex\left\vert #1 
    \right\vert\kern-0.25ex\right\vert\kern-0.25ex\right\vert}}
\newcommand{\floor}[1]{\left\lfloor#1\right\rfloor}
\newcommand{\ceil}[1]{\left\lceil#1\right\rceil}
\newcommand{\overbar}[1]{\mkern 1mu \overline{\mkern-1.5mu#1\mkern0mu}\mkern 1mu}
\newcommand{\E}{\mathbb{E}}

\newcommand{\N}{\mathbb{N}}

\newcommand{\R}{\mathbb{R}}

\newcommand{\CC}{\mathcal{C}}

\newcommand{\GG}{\mathcal{G}}

\newcommand{\NN}{\mathcal{N}}

\newcommand{\OO}{\mathcal{O}}

\newcommand{\dd}{\mathrm{d}}

\newcommand{\one}{\mathds{1}}

\newcommand{\lr}{\eta}

\newcommand{\relu}{\mathrm{ReLU}}
\newcommand{\sign}{\mathrm{sign}}

\newcommand{\iid}{\overset{{\tiny \text{i.i.d}}}{\sim}}
\newcommand{\tR}{t\wedge\theta_R}
\newcommand{\tauR}{\tau\wedge\theta_R}

\newcommand{\hbarr}{\overbar{H}}
\newcommand{\htilde}{\widetilde{H}}
\newcommand{\Xtilde}{\widetilde{X}}
\newcommand{\hhat}{\widehat{h}}

\newcommand{\qedwhite}{\begin{flushright} \ensuremath{\Box} \end{flushright}}

\newtheorem{theorem}{Theorem}[section]

\newtheorem{lemma}[theorem]{Lemma}
\newtheorem{remark}[theorem]{Remark}
\newtheorem{assumption}[theorem]{Assumption}
\newtheorem{hypothesis}{Hypothesis}

\newtheorem{proposition}[theorem]{Proposition}

%% file: sec/1_introduction.tex
\section{Introduction} \label{sec:intro}


Residual networks, or ResNets, are multilayer neural network architectures in which a {\it skip connection} is introduced at every layer~\cite{HZRS2016}. This allows  deep networks to be trained by circumventing vanishing and exploding gradients~\cite{BSF1994}. The increased depth in ResNets has lead to commensurate performance gains in applications ranging from speech recognition~\cite{HDH2016, ZK2016} to computer vision~\cite{HZRS2016, HSLSW2016}. 

A residual network with $L$ layers may be represented as
\begin{equation}\label{forward-map-resnet}
    h^{(L)}_{k+1} = h^{(L)}_{k} + \delta^{(L)}_k \sigma_d\left(A^{(L)}_{k}h^{(L)}_{k} + b^{(L)}_k \right),
\end{equation} 
where $h^{(L)}_k$ is the hidden state at layer $k=0,\ldots,L$, $h^{(L)}_0 = x \in \R^d$ the input, $h^{(L)}_L \in \R^d$ the output, $\sigma\colon\mathbb{R}\to\mathbb{R}$ is a nonlinear activation function, $\sigma_d(x) = (\sigma(x_1),\ldots,\sigma(x_d))^{\top}$ its component-wise extension to $x\in \mathbb{R}^d$, and $A_k^{(L)}$, $b_k^{(L)}$, and $\delta_k^{(L)}$ are trainable network weights for $k=0, \ldots, L-1$.


\subsection{Connection to previous work}

ResNets have been the focus of several theoretical studies due to a perceived link with a class of differential equations. The idea, put forth in~\cite{HR2018,CRBD2018}, is to view \eqref{forward-map-resnet} as a discretization of a system of ordinary differential equations 
\begin{equation}\label{eq:neural_ode_limit}
    \frac{\dd H_t}{\dd t}= \sigma_d\left(\overbar{A}_t H_t + \overline{b}_t \right),
\end{equation}
where $\overbar{A}\colon[0,1]\to\R^{d\times d}$ and $\overline{b}\colon[0,1]\to\R^d$ are appropriate smooth functions and $H(0)=x$. 
This may be justified~\cite{TvG2018} by assuming that 
\begin{equation}
    \delta^{(L)}\sim 1/L,\quad A_k^{(L)}\to \overbar{A}_{t},\quad b_k^{(L)}\to \overline{b}_{t} \label{eq.ThorpeScaling}
\end{equation}
as $L$ increases and $k/L \to t$. Such models, named neural ordinary differential equations or neural ODEs~\cite{CRBD2018,dupont2019augmented}, have motivated the use of optimal control methods to train ResNets~\cite{ew2018}.


However, the precise link between deep ResNets and the neural ODE model~\eqref{eq:neural_ode_limit} is unclear: in practice, the weights $A^{(L)}$ and $b^{(L)}$ result from training, and the validity of the scaling assumptions~\eqref{eq.ThorpeScaling} for trained weights  is far from obvious and has not been verified.
As a matter of fact, there is empirical evidence showing that using a scaling factor $\delta^{(L)}\sim 1/L$ can deteriorate the network accuracy~\cite{BMMCM2020}. Also, there is no guarantee that weights obtained through training have a non-zero limit which depends smoothly on the layer, as~\eqref{eq.ThorpeScaling} would require. In fact, we present numerical experiments which point to the contrary for many ResNet architectures used in practice. These observations motivate an in-depth examination of the actual scaling behavior of weights with network depth in ResNets and of its impact on the asymptotic behavior of those networks.




%

\begin{figure*}[h!]
    \centering
    \includegraphics[width=0.32\textwidth]{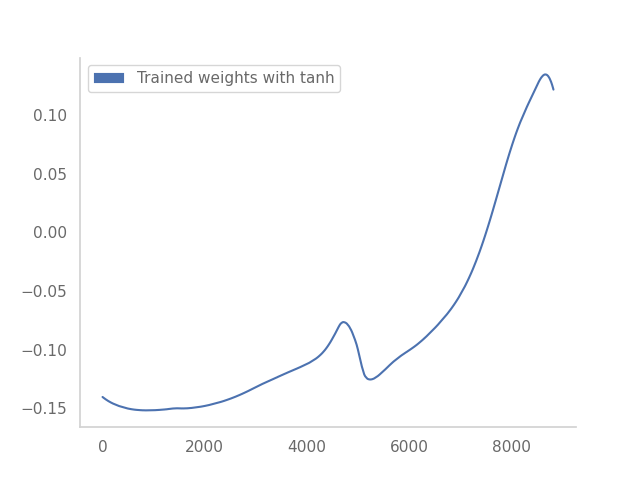}
    \includegraphics[width=0.32\textwidth]{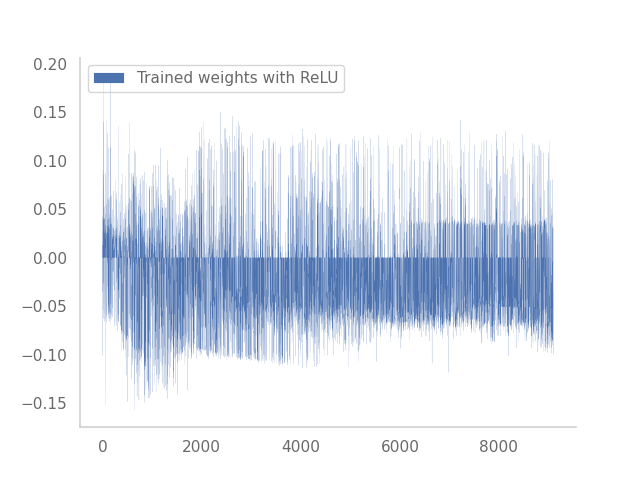}
    \includegraphics[width=0.32\textwidth]{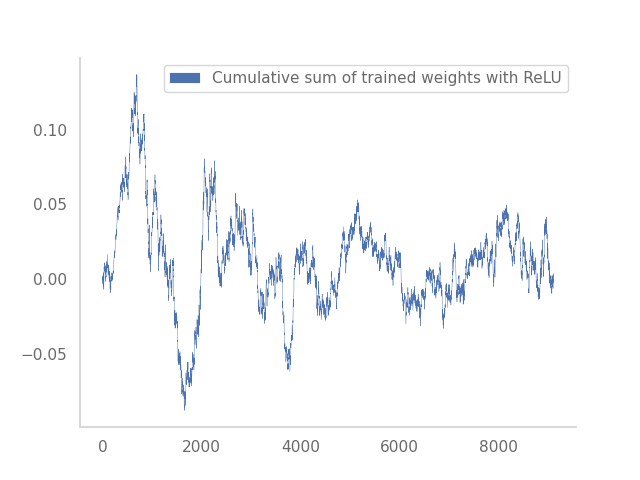}
    \caption{Trained weights as a function of $k$ for $k=0,\ldots,L$ and $L=9100$. {\bf Left}: rescaled weights $L^\beta A^{(L)}_{k, (0, 0)}$ for a $\tanh$ network with $\beta=0.2$. {\bf Center}: weights $A^{(L)}_{k, (0, 0)}$ for a $\relu$ network.  {\bf Right}: cumulative sum  $\sum_{j=0}^{k-1} A^{(L)}_{j, (0, 0)}$for a $\relu$ network. 
    \label{fig:trained_weights}}
\end{figure*}

\subsection{Our contributions}

We systematically investigate the scaling behavior of trained ResNet weights as the number of layers increases and examine the consequences of this behavior for the asymptotic properties of deep ResNets. Our code is publicly available at \url{https://github.com/instadeepai/scaling-resnets}.

Our main contributions are twofold. 
Using the methodology described in Section~\ref{sec:methodology}, we design detailed numerical experiments to study the scaling of trained network weights across a range of ResNet architectures and datasets, showing the existence of at least three different scaling regimes, none of which correspond to~\eqref{eq.ThorpeScaling}. In Section~\ref{sec:results}, we show that in two of these scaling regimes, the properties of deep ResNets may be described in terms of a class of ordinary or stochastic differential equations, albeit different from the neural ODEs studied in~\cite{CRBD2018,HR2018, lu2020mean}. Those novel findings on the relation between ResNets and differential equations complement previous work~\cite{TvG2018,DBLP:journals/corr/abs-1904-05263,DBLP:journals/corr/abs-1910-02934,ott2021resnet}.
In particular, our findings question the validity of the neural ODE \eqref{eq:neural_ode_limit} as a description of deep ResNets with trained weights.

\subsection{Notations}
 $\|y\| $ denotes the Euclidean norm of a vector $y$. For a matrix $x$,  $x^{\top}$ denotes its transpose, ${\rm Diag}(x)$ its diagonal vector, ${\rm Tr}(x)$ its trace and $\norm{x}_F = \sqrt{{\rm Tr}(x^{\top}x)}$ its Frobenius norm.  $\floor{u}$ denotes the integer part of a positive number $u$. 
 $\NN(m, v)$ denotes the Gaussian distribution with mean $m$ and (co)variance $v$, $\otimes$ denotes the tensor product, and $\R^{d, \, \otimes n} = \R^d \times \cdots \times \R^d$ ($n$ times).  ${\rm vec} \colon \R^{d_1{\times\cdots\times} d_n} \to \R^{d_1\cdots d_n}$ denotes the  vectorisation operator, and  $\one_S$  the indicator function of a set $S$. $\CC^0$ is the space of continuous functions, and for $\nu\geq0$, $\CC^\nu$ is the space of $\nu$-Hölder continuous functions.

%% file: sec/2_methodology.tex
\section{Methodology} \label{sec:methodology}
We identify the possible scaling regimes for the network weights, introduce the quantities needed to characterize the deep network limit, and describe the step-by-step procedure we use to analyze our numerical experiments.

\subsection{Scaling hypotheses}

As described in Section~\ref{sec:intro}, the neural ODE limit assumes  
\begin{equation} \label{eq:thorpe}
\delta^{(L)} \sim  \frac{1}{L} \quad {\rm and} \quad A^{(L)}_{\floor{Lt} }\overset{L\to \infty}{\longrightarrow} \overbar{A}_t,\quad b^{(L)}_{\floor{Lt} }\overset{L\to \infty}{\longrightarrow} \overbar{b}_t,
\end{equation}
for $t\in [0,1]$ where $\overbar{A} \colon [0,1] \to \R^{d\times d}$ and $\overbar{b} \colon [0,1] \to \R^{d\times d}$ are   smooth  functions ~\cite{TvG2018}.  
Our numerical experiments, detailed in Section~\ref{sec:experiments}, show that   the weights generally shrink as $L$ increases (see for example Figures~\ref{fig:scaling_tanh_shared_delta} and~\ref{fig:scaling_relu_scalar_delta}), so one cannot expect the above assumption to hold, and weights need to be renormalized in order to converge to a non-zero limit.
We consider here the following more general situation which includes~\eqref{eq:thorpe} but allows for shrinking weights: 
\begin{hypothesis} \label{hypothesis.1}
There exist $\overbar{A} \in \CC^0\left( [0,1], \R^{d\times d} \right)$ and $\beta \in \left[0, 1\right]$ such that 
\begin{equation} \label{scaling-beta}
\forall s\in [0,1],  \qquad   \overbar{A}_s = \lim_{L\to\infty} L^{\beta}\,  A^{(L)}_{\floor{Ls}}.
\end{equation}
\end{hypothesis}
Properly renormalized weights may indeed converge to a continuous function of the layer in some cases, as shown in Figure~\ref{fig:trained_weights} (left) which displays an example of layer dependence of trained weights for a ResNet~\eqref{forward-map-resnet} with fully connected layers and $\tanh$ activation function, without explicit regularization (see Section \ref{sec:fully_connected}).


However it is not always the case that network weights  converge to a smooth function of the layer, even after rescaling. For example, network weights $A_k^{(L)}$ are usually initialized to random, independent and identically distributed (i.i.d.) values, whose scaling limit would then correspond to a {\it white noise}, which cannot be represented as a function of the layer. Such scaling behaviour also occurs for trained weights, as shown in Figure \ref{fig:trained_weights} (center). In this case, the {\it cumulative sum} $\sum_{j=0}^{k-1} A^{(L)}_j$ of the weights behaves like a random walk, which does have a well-defined scaling limit $W \in \CC^0\left(\left[0, 1\right], \R^{d\times d} \right)$.
Figure \ref{fig:trained_weights} (right) shows that, for a $\relu$ ResNet with fully-connected layers, this cumulative sum of trained weights converges to an {\it irregular}, that is, non-smooth function of the layer. 

This observation motivates the consideration of an alternative hypothesis where the weights $A^{(L)}_k$ are represented as the {\it increments} of a continuous function $W^A$. 

Combining such terms with the ones considered in Hypothesis \ref{hypothesis.1}, we consider the following, more general, setting:
\begin{hypothesis} \label{hypothesis.2}
There exist $\beta \in \left[0, 1\right)$, $\overbar{A} \in \CC^0\left( [0,1], \R^{d\times d} \right)$, and $W^A \in \CC^{0}([0,1], \mathbb{R}^{d\times d} )$ non-zero such that $W^A_0=0$ and
\begin{equation} \label{scaling-W}
A_k^{(L)} = L^{-\beta} \overbar{A}_{k/L} +  W^A_{(k+1)/L} - W^A_{k/L}.
\end{equation}
\end{hypothesis}
The above decomposition is unique. Indeed, for $s\in[0,1]$,
\begin{align}
L^{\beta -1} \sum_{k=0}^{\floor{Ls}-1} A_k^{(L)} &= L^{-1} \sum_{k=0}^{\floor{Ls}-1} \overbar{A}_{k/L} + L^{\beta - 1} W^A_{\floor{Ls}/L} \nonumber \\
&\to \int_{0}^s \overbar{A}_r \dd r, \quad \mbox{as}\, L \to \infty. \label{cumul-decomp}
\end{align}

The integral of $\overbar{A}$ is thus uniquely determined by the weights $A_k^{(L)}$ when $L$ is large, so $\overbar{A}$ can be obtained by discretization and $W^A$ by fitting the residual error in \eqref{cumul-decomp}. In addition, Hypotheses~\ref{hypothesis.1} and~\ref{hypothesis.2} are mutually exclusive since Hypothesis~\ref{hypothesis.2} requires $W^A$ to be non-zero.

\begin{remark}[IID initialization of weights] \label{rem:weight-init}
In the special case of independent Gaussian weights 
\begin{equation*}
    A_{k,mn}^{(L)} \iid \mathcal{N}\left(0, L^{-1} d^{-2} \right) \quad \text{and} \quad  b_{k,n}^{(L)} \iid \mathcal{N}\left(0, L^{-1}d^{-1}\right)
\end{equation*}
where $A_{k,mn}^{(L)}$ is the $(m,n)$-th entry of $A_k^{(L)}\in\R^{d\times d}$ and $b_{k,n}^{(L)}$ is the $n$-th entry of $b_k^{(L)}\in\R^d$, 
 we can represent the weights $\{ A^{(L)}, b^{(L)}\}$ as the increments of a matrix Brownian motion
\begin{equation*}
    A_{k}^{(L)} = d^{-1}\left(W^A_{(k+1)/L} - W^A_{k/L}\right),
\end{equation*}
which is a special case of Hypothesis 2.
\end{remark}



\subsection{Smoothness of weights with respect to the layer} \label{sec:tools}

A question related to the existence of a scaling limit is the degree of smoothness of the limits $\overbar{A}$ or $W^A$, if they exist. 
To quantify the smoothness of the function mapping the layer number to the corresponding network weight, we define in Table~\ref{tab:summary_results} several quantities which may be viewed as discrete versions of various (semi-)norms used to measure the smoothness of functions.

{
    \tabulinesep=1.3mm
        \captionof{table}{Quantities associated to a tensor $A^{(L)} \in \R^{L\times d \times d}$.}
    \begin{center}
    \begin{tabu}{p{3.2cm}|p{3.9cm}}
        \hline
        \textbf{Quantity} & \textbf{Definition}  \\ \hline
        Maximum norm & $\max_{k} \norm{A^{(L)}_k}_F$ \\
        $\beta$-scaled norm \newline of increments & $L^{\beta} \max_{k} \norm{A^{(L)}_{k+1} - A^{(L)}_{k}}_F$ \\
        Cumulative sum norm & $\norm{ \sum_{k} A_k^{(L)} }_F$ \\
        Root sum of squares & $\left(\sum_k \norm{A^{(L)}_k}_F^2\right)^{1/2}$ \\
        \hline
    \end{tabu}
    \end{center}
    \label{tab:summary_results}
}

The first two norms relate to Hypothesis 1: if $A^{(L)}$ satisfy \eqref{scaling-beta}, then the maximum norm scales like $L^{-\beta}$ and the $\beta$-scaled norm of increments scales like $L^{-\nu}$ if the limit function $\overbar{A}$ is $\nu$-Hölder-continuous. \\
The last two norms relate to Hypothesis 2: if $A^{(L)}$ satisfy \eqref{scaling-W}, then the cumulative sum norm scales like $L^{-\beta}$. Furthermore, the root sum of squares gives us the regularity of $W^A$. Indeed, define the \textit{quadratic variation tensor} of $W^A$ by
\begin{align*}
\left[ W^A \right]_s &= \lim_{L \to \infty} \sum_{k=0}^{\floor{Ls}-1} \left( W^A_{\frac{k+1}{L}} - W^A_{\frac{k}{L}} \right) \otimes  \left( W^A_{\frac{k+1}{L}} - W^A_{\frac{k}{L}} \right)^{\top}.
\end{align*}
Then, using~\eqref{scaling-W} and Cauchy-Schwarz, we estimate 
\begin{equation} \label{qv-sqrtssq}
\tensornorm{\left[ W^A \right]_s} \leq 2 \cdot \lim_{L\to\infty} \sum_{k=0}^{\floor{Ls}-1} \norm{A^{(L)}_k}_F^2 + L^{1-2\beta} \norm{\overbar{A}}^2_{L^2}
\end{equation}
where $\tensornorm{\cdot}$ is the Hilbert-Schmidt norm. As $\overbar{A}$ is continuous on a compact domain, its $L^2$ norm is finite. Hence, if $\beta \geq 1/2$, the fact that the root sum of squares of $A^{(L)}$ is upper bounded as $L\to\infty$ implies that the quadratic variation of $W^A$ is finite.

\subsection{Procedure for  numerical experiments} \label{sec:procedure-experiments}

Note that Hypotheses~\ref{hypothesis.1} and~\ref{hypothesis.2} are mutually exclusive since Hypothesis~\ref{hypothesis.2} requires $W^A$ to be non-zero. In order to examine whether one of these hypotheses, or neither, holds for the trained weights $A^{(L)}$ and $b^{(L)}$, we proceed as follows.


\underline{Step 1}: We perform a logarithmic regression of the maximum norm of $\delta^{(L)}$ with respect to $L$ to deduce the scaling $\delta^{(L)} \sim L^{-\alpha}$.

\underline{Step 2}: To obtain the exponent $\beta\in [0,1)$, we perform a logarithmic regression of the cumulative sum norm of $A^{(L)}$ with respect to $L$. Indeed, \eqref{cumul-decomp} for $s=1$ indicates that the cumulative sum norm explodes with a slope of $1-\beta$.

\underline{Step 3}: After identifying the correct exponent $\beta$ for the weights, we compute the $\beta$-scaled norm of increments of $A^{(L)}$ to check Hypothesis~\ref{hypothesis.1} and measure the smoothness of the trained weights. On one hand, if the $\beta$-scaled norm of increments of $A^{(L)}$ does not vanish as $L\to\infty$, it means that the rescaled weights cannot be represented as a continuous function of the layer, as in Hypothesis~\ref{hypothesis.1}. On the other hand, if the $\beta$-scaled norm of increments of $A^{(L)}$ vanishes (say, as $L^{-\nu}$) when $L$ increases, it supports Hypothesis~\ref{hypothesis.1} with a Hölder-continuous limit function $\overbar{A}\in \CC^\nu([0,1], \R^{d\times d})$.

\underline{Step 4}: To discriminate between Hypothesis~\ref{hypothesis.1} and Hypothesis~\ref{hypothesis.2}, we decompose the cumulative sum $\sum_{j=0}^{k-1} A_j^{(L)}$ of the trained weights into a {\it trend} component $\overbar{A}$ and a {\it noise} component $W^A$, as shown in~\eqref{cumul-decomp}. The presence of non-negligible noise term $W^A$ favors Hypothesis~\ref{hypothesis.2}.

\underline{Step 5}: Finally, we estimate the regularity of the term $W^A$ under Hypothesis~\ref{hypothesis.2}. If $\beta\geq 1/2$ and the root sum of squares of $A^{(L)}$ is finite, we deduce by \eqref{qv-sqrtssq} that $W^A$ is of finite quadratic variation. It happens for example when $W^A$ has a {\it diffusive} behavior, as in the example of i.i.d. random weights.

The same procedure is used for $b^{(L)}$. Note that the scaling exponent $\beta$ may be different for $A^{(L)}$ and $b^{(L)}$.

\begin{remark}\label{rmk:relu}
Note that $\sigma = \relu$ is homogeneous of degree $1$, so we can write
\begin{equation*}
\delta \cdot \sigma_d\left(Ah+b \right) = \sign(\delta) \cdot \sigma_d\left(\abs{\delta}A h + \abs{\delta}b \right).
\end{equation*}
Hence, when analyzing the scaling of trained weights in the case of a $\relu$ activation with fully-connected layers, we look at the quantities $\abs{\delta^{(L)}} A^{(L)}$ and $\abs{\delta^{(L)}} b^{(L)}$, as they represent the total scaling of the residual connection. \end{remark}

%% file: sec/3_experiments.tex
\section{Numerical Experiments} \label{sec:experiments}

We investigate the scaling properties and asymptotic behavior of trained weights for residual networks as the number of layers increases. We focus on two types of architectures: fully-connected and convolutional networks.

\subsection{Fully-connected layers \label{sec:fully_connected}}

\paragraph{Architecture.}
We consider a regression problem where the network layers are fully-connected. We consider the network architecture~\eqref{forward-map-resnet} for two different setups:
\begin{enumerate}[(i)]
    \item $\sigma = \tanh$, $\delta^{(L)}_k = \delta^{(L)} \in \R_+$ trainable,
    \item $\sigma = \relu$, $\delta^{(L)}_k \in \R$ trainable.
\end{enumerate}
We choose to present these two cases for the following reasons. First, both $\tanh$ and $\relu$ are widely used in practice. Further, having $\delta^{(L)}$ scalar makes the derivation of the limiting behavior simpler. Also, since $\tanh$ is an odd function, the sign of $\delta^{(L)}$ can be absorbed into the activation. Therefore, we can assume that $\delta^{(L)}$ is non-negative for $\tanh$. Regarding $\relu$, having a shared $\delta^{(L)}$ would hinder the expressiveness of the network. Indeed, if for instance $\delta^{(L)} > 0$, we would get $h^{(L)}_{k+1} \geq  h^{(L)}_{k}$ element-wise since $\relu$ is non-negative. This would imply that $h^{(L)}_L \geq x$, which is not desirable. The same argument applies to the case $\delta^{(L)} < 0$. Thus, we let $\delta^{(L)}_k \in \R$ depend on the layer number for $\relu$ networks.

\paragraph{Datasets and training.}
We consider two datasets. The first one is synthetic: fix $d=10$ and generate $N$ i.i.d samples $x_i$ coming from the $d-$dimensional uniform distribution in $[-1, 1]^d$. Let $K=100$ and simulate the following dynamical system:
\begin{equation*}
    \begin{cases}
    z_{0}^{x_i}&=x_i \\
    z_{k}^{x_i} &= z_{k-1}^{x_i}+ K^{-1/2}\tanh_d\left(g_d\left(z_{k-1}^{x_i},k, K\right) \right),
    \end{cases}
\end{equation*}
where 
$g_d(z,k,K)\coloneqq\sin(5k\pi/K)z+\cos( 5k\pi/K )\one_d$. The targets $y_i$ are defined as $y_i = z_{K}^{x_i} / \norm{z_{K}^{x_i}}$. The motivation behind this low-dimensional dataset is to be able to train very deep residual networks on a problem where the optimal input-output map lies inside the class of functions represented by \eqref{forward-map-resnet}.

The second dataset is a low-dimensional embedding of the MNIST handwritten digits dataset~\cite{MNIST}. Let $\left( \widetilde{x}, c \right) \in \R^{28\times 28} \times \left\{0, \ldots, 9 \right\}$ be an input image and its corresponding class. We transform $\widetilde{x}$ into a lower dimensional embedding $x \in \R^{d}$ using an untrained convolutional projection, where $d=25$. More precisely, we stack two convolutional layers initialized randomly, we apply them to the input and we flatten the downsized image into a $d-$dimensional vector. Doing so reduces the dimensionality of the problem while allowing very deep networks to reach at least $99\%$ training accuracy. The target $y\in \R^d$ is the one-hot encoding of the corresponding class. 

The weights are updated by stochastic gradient descent (SGD)  on the unregularized mean-squared loss using batches of size $B$ and a constant learning rate $\lr$. We perform SGD updates until the loss falls below $\epsilon$, or when the maximum number of updates $T_{\max}$ is reached. We repeat the experiments for several depths $L$ varying from $L_{\min}$ to $L_{\max}$. All the hyperparameters are given in Appendix~\ref{sec:hyperparameters}.

\paragraph{Results.}

\begin{figure}[h!]
    \centering
    \includegraphics[width=0.238\textwidth]{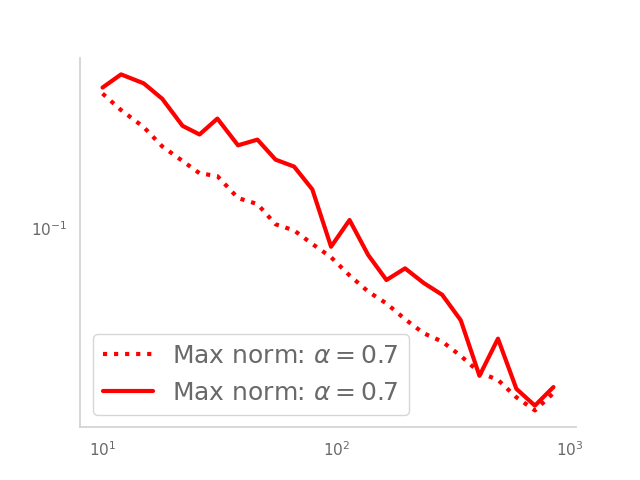}
    \includegraphics[width=0.238\textwidth]{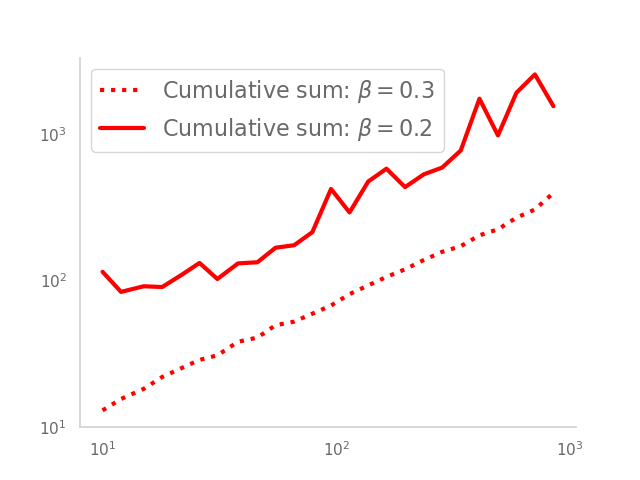}
    \caption{Scaling for $\tanh$ activation and $\delta^{(L)}\in\R$. Left: Maximum norm of $\delta^{(L)}$ with respect to $L$. Right: Cumulative sum norm of $A^{(L)}$ with respect to $L$. The dashed lines are for the synthetic data and the solid lines are for MNIST. The plots are in log-log scale.}
    \label{fig:scaling_tanh_shared_delta}
\end{figure}

For the case of a $\tanh$ activation (i), we observe in Figure~\ref{fig:scaling_tanh_shared_delta} that for both datasets, $\delta^{(L)} \sim L^{-0.7}$ clearly decreases as $L$ increases, and the cumulative sum norm of $A^{(L)}$ slightly increases when $L$ increases. We deduce that $\beta = 0.2$ for the MNIST dataset and $\beta = 0.3$ for the synthetic dataset.

\begin{figure}[h!]
    \centering
    \includegraphics[width=0.238\textwidth]{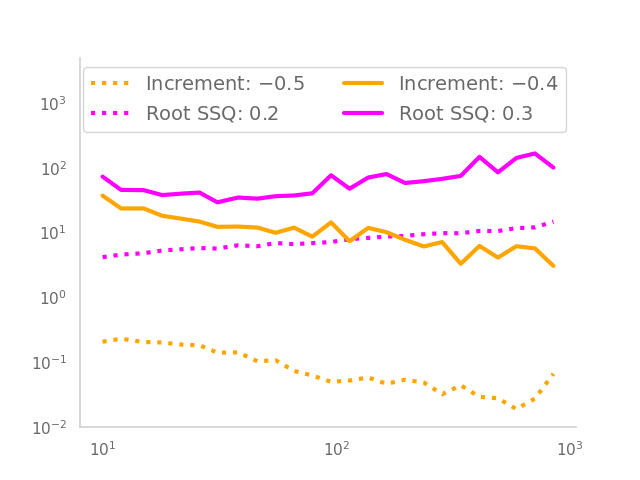}
    \includegraphics[width=0.238\textwidth]{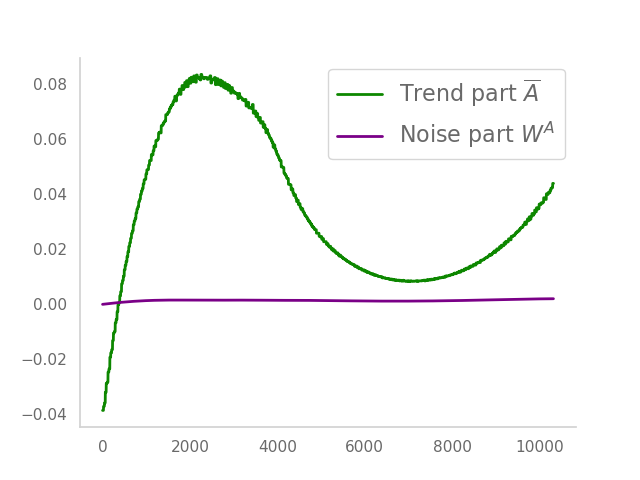}
    \caption{Hypothesis verification for $\tanh$ activation and $\delta^{(L)}\in\R$. Left: in pink we plot in log-log scale the root sum of squares of $A^{(L)}$, and in orange the $\beta$-scaled norm of increments of $A^{(L)}$. The dashed lines are for the synthetic data and the solid lines are for MNIST. Right: Decomposition of the trained weights $A^{(L)}_{k, (9, 7)}$ with the trend part $\overbar{A}$ and the noise part $W^A$ for $L=10321$, as defined in~(\ref{scaling-W}), for the synthetic dataset.}
    \label{fig:hypothesis_tanh_shared_delta}
\end{figure}

Next, we verify which of Hypothesis~\ref{hypothesis.1} or Hypothesis~\ref{hypothesis.2} holds for $A^{(L)}$. We observe in Figure~\ref{fig:hypothesis_tanh_shared_delta} (left) that the $\beta$-scaled norm of increments of $A^{(L)}$ decreases like $L^{-1/2}$, suggesting that Hypothesis~\ref{hypothesis.1} holds, with $\overbar{A}$ being $1/2-$Hölder continuous. This is confirmed in Figure~\ref{fig:hypothesis_tanh_shared_delta} (right), as the trend part $\overbar{A}$ is visibly continuous and even of class $\CC^1$. The noise part $W^A$ is negligible. This observation is even more striking given that the weights are trained \textbf{without explicit regularization}.

\begin{figure}[h!]
    \centering
    \includegraphics[width=0.238\textwidth]{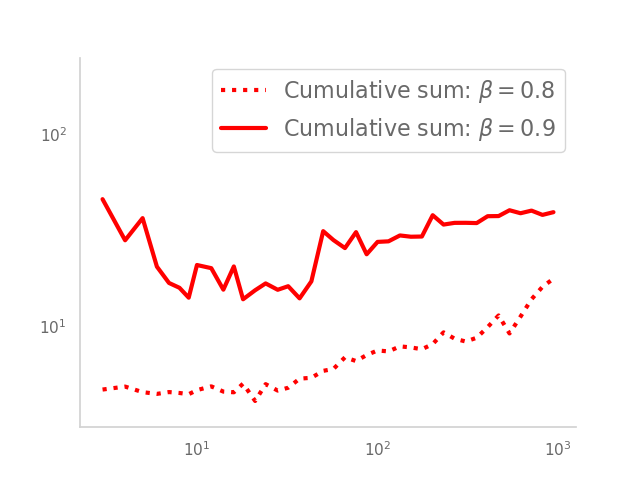}
    \includegraphics[width=0.238\textwidth]{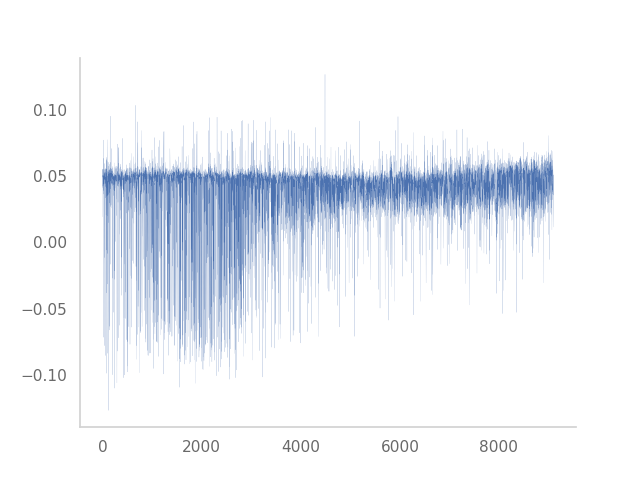}
    \caption{Scaling for $\relu$ activation and $\delta_k^{(L)}\in\R$. Left: Cumulative sum norm of $\vert \delta^{(L)}\vert A^{(L)}$ with respect to $L$, in log-log scale. Right: trained values of $\delta^{(L)}_k$ as a function of $k$, for $L=9100$ and for the synthetic dataset.}
    \label{fig:scaling_relu_scalar_delta}
\end{figure}

Regarding the case of a $\relu$ activation function (ii), we observe in Figure~\ref{fig:scaling_relu_scalar_delta} (left) that the cumulative sum norm of the residual connection $\abs{ \delta^{(L)}} A^{(L)}$ scales like $L^{0.2}$ for the synthetic dataset and like $L^{0.1}$ for the MNIST dataset, so $\beta=0.8$, resp. $0.9$ in this case. We see in Figure~\ref{fig:scaling_relu_scalar_delta} (right) that keeping the sign of $\delta^{(L)}_k$ is important, as the sign oscillates considerably throughout the network depth $k=0, \ldots, L-1$.

\begin{figure}[h!]
    \centering
    \includegraphics[width=0.238\textwidth]{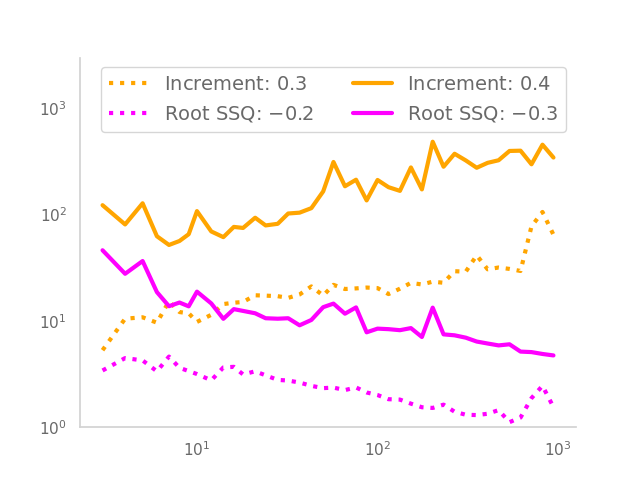}
    \includegraphics[width=0.238\textwidth]{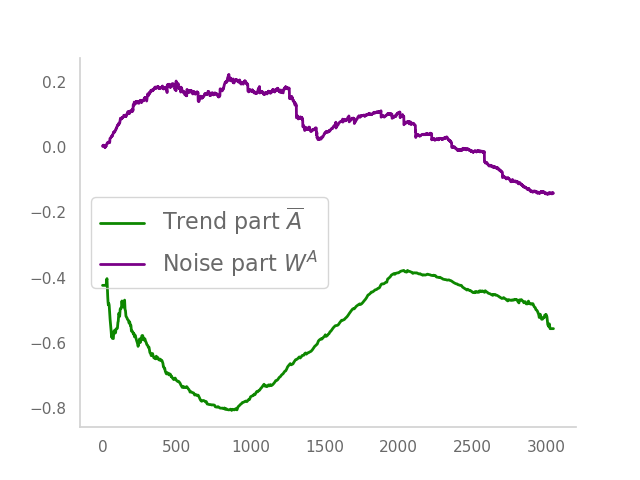}
    \caption{Hypothesis verification for $\relu$ activation and $\delta_k^{(L)}\in\R$. Left: in pink we plot in log-log scale the root sum of squares of $\vert \delta^{(L)}\vert A^{(L)}$, and in orange the $\beta$-scaled norm of increments of $\vert \delta^{(L)}\vert A^{(L)}$. The dashed lines are for the synthetic data and the solid lines for MNIST. Right: Decomposition of the trained weights $ \vert \delta^{(L)} \vert \, A^{(L)}_{k, (7, 7)}$ with the trend part $\overbar{A}$ and the noise part $W^A$ for $L=10321$, as defined in~(\ref{scaling-W}), for the synthetic dataset.
    \label{fig:hypothesis_relu_delta_scalar}}
\end{figure}

We verify now which of Hypothesis~\ref{hypothesis.1} or Hypothesis~\ref{hypothesis.2} holds for $\abs{\delta^{(L)}} A^{(L)}$. Figure~\ref{fig:hypothesis_relu_delta_scalar} (left) shows that the $\beta$-scaled norm of increments scales like $L^{0.2}$ and $L^{0.4}$ as the depth increases. This suggests that there exists a noise part $W^A$. Following~\eqref{qv-sqrtssq}, the fact that the root sum of squares of $\abs{\delta^{(L)}} A^{(L)}$ is upper bounded as $L\to\infty$ implies that $W^A$ has finite quadratic variation. These claims are also supported by Figure~\ref{fig:hypothesis_relu_delta_scalar} (right): there is a non-zero trend part $\overbar{A}$, and a non-negligible noise part $W^A$.

Given the scaling behavior of the trained weights, we conclude that Hypothesis~\ref{hypothesis.1} seems to be a plausible description for the $\tanh$ case (i), but Hypothesis~\ref{hypothesis.2} provides a better description for the $\relu$ case (ii).

The same conclusions hold for $b^{(L)}$ as well, see Appendix~\ref{app:further_plots}.

\paragraph{Role of the noise term $W^A$.}

\begin{figure}[h!]
    \centering
    \includegraphics[width=0.238\textwidth]{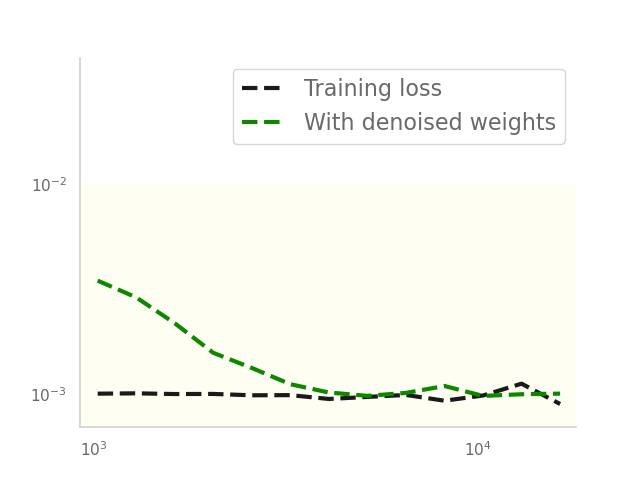}
    \includegraphics[width=0.238\textwidth]{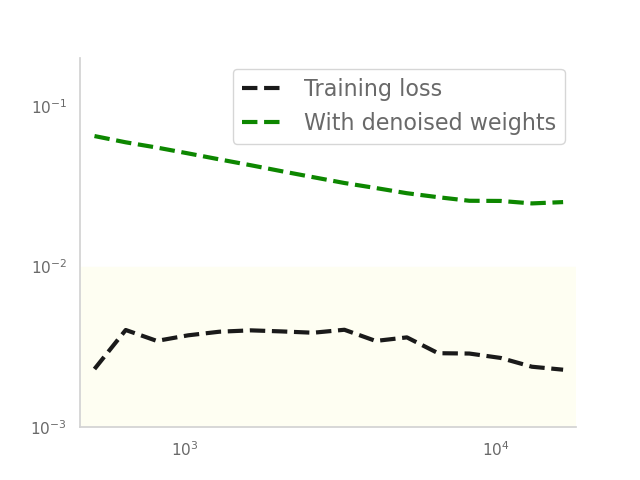}
    \caption{Loss value, as a function of $L$, in black for the trained weights $A_k^{(L)}$ and in green for the denoised weights $\widetilde{A}^{(L)}_k = L^{-\beta} \overbar{A}_{k/L}$. Left: $\tanh$ activation and $\delta^{(L)}\in\R$. Right: $\relu$ activation and $\delta_k^{(L)}\in\R$. Note that these curves are for the synthetic dataset and that we plot them in log-log scale. Also, we show in off-white the loss value range in which we consider our networks to have converged.}
    \label{fig:loss_denoised}
\end{figure}

A legitimate question to ask at this point is whether the noise part $W^A$ plays a significant role in the accuracy of the network. To test this, we create a residual network with denoised weights $\widetilde{A}^{(L)}_k \coloneqq  L^{-\beta} \overbar{A}_{k/L}$, compute its training error and we compare it to the original training error. We observe in Figure~\ref{fig:loss_denoised} (left) that for $\tanh$, the noise part $W^A$ is negligible and does not influence the loss. However, for $\relu$, the loss with denoised weights is one order of magnitude above the original training loss, meaning that the noise part $W^A$ plays a significant role in the accuracy of the trained network.

\subsection{Convolutional layers \label{sec:classification}}

We now consider the original ResNet with convolutional layers introduced in~\cite{HZRS2016}. This architecture is close to the state-of-the-art methods used for image recognition. In particular, we do not include batch normalization~\cite{DBLP:journals/corr/IoffeS15} since it only slightly improves the performance of the network while making the analysis significantly more complicated.

\paragraph{Setup.}
The precise architecture is detailed in Appendix~\ref{sec:resnet_architectures}. Most importantly, our network still possesses the key skip connections from~(\ref{forward-map-resnet}). Simply, the update rule for the hidden state reads
\begin{equation} \label{eq:2d-forward-map}
    h_{k+1} = \sigma\left(h_{k} +\Delta_k * \sigma\left(A_k*h_{k}\right)+F_k*h_{k}\right)
\end{equation}
for $k=0,\ldots,L-1$, where $\sigma=\relu$. Here, $\Delta_k$, $A_k$, and $F_k$ are kernels and $*$ denotes convolution. Note that $\Delta_k$ plays the same role as $\delta_k^{(L)}$ from~(\ref{forward-map-resnet}). To lighten the notation, we omit the superscripts $x$ (the input) and $L$ (the number of layers).

We train our residual networks at depths ranging from $L_{\min}=8$ to $L_{\max}=121$ on the CIFAR-10~\cite{CIFAR-10} dataset with the unregularized cross-entropy loss. Here, the depth is the number of residual connections. We underline that a network with $L_{\max}=121$ is already very deep. As a comparison, a standard ResNet-152~\cite{HZRS2016} has depth $L=50$ in our framework.


\begin{figure}[h!]
    \includegraphics[width=.238\textwidth]{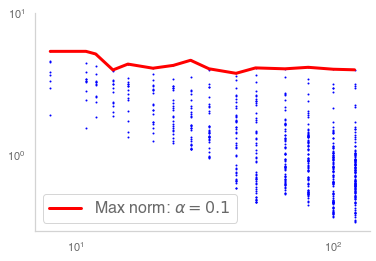}
    \includegraphics[width=.238\textwidth]{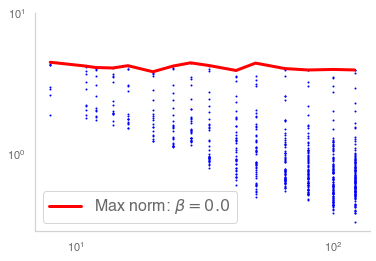}
    \caption{Scaling of $\Delta^{(L)}$ (left) and $A^{(L)}$ (right) against the network depth $L$ for convolutional architectures on CIFAR-10. In blue, we plot the spectral norm of the kernels $\Delta^{(L)}_k$, resp. $A^{(L)}_k$, for $k=0, \ldots, L-1$. The red line is the maximum of these values over $k$, namely the maximum norm, defined in Table~\ref{tab:summary_results}. The plots are in log-log scale.}
    \label{fig:cifar_scaling_with_depth}
\end{figure}
\paragraph{Results.}

As in Section~\ref{sec:fully_connected}, we investigate how the weights scale with network depth and whether Hypothesis~\ref{hypothesis.1} or Hypothesis~\ref{hypothesis.2} holds  for a convolutional networks. To that end, we compute  the spectral norms, of the linear operators defined by the convolutional kernels $\Delta_k^{(L)}$ and $A_k^{(L)}$ using the method described in~\cite{DBLP:journals/corr/abs-1805-10408}. Figure~\ref{fig:cifar_scaling_with_depth} shows the maximum norm, and hence the scaling of $\Delta^{(L)}$ and $A^{(L)}$ against the network depth $L$. We observe that $\Delta^{(L)} \sim L^{-\alpha}$ and $A^{(L)} \sim L^{-\beta}$ with $\alpha=0.1$ and $\beta = 0$.



\begin{figure}[h!]
    \includegraphics[width=.238\textwidth]{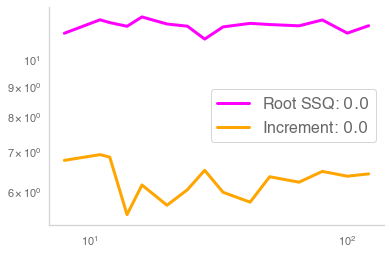}
    \includegraphics[width=.238\textwidth]{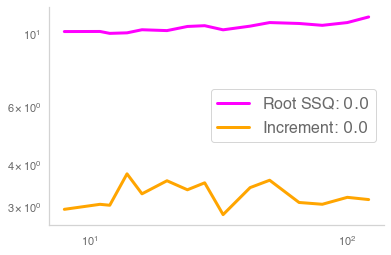}
    \caption{Testing of Hypotheses~\ref{hypothesis.1} and~\ref{hypothesis.2} for $\Delta^{(L)}$ (left) and $A^{(L)}$ (right). We plot in pink the root sum of squares and in orange the $\alpha$-scaled norm of increments of $\Delta^{(L)}$ (left) and the $\beta$-scaled norm of increments of $A^{(L)}$ (right). Plots are in log-log scale. The root sum of squares and the scaled norm of increments are defined in Table~\ref{tab:summary_results}. We obtain $\alpha$ and $\beta$ from Figure~\ref{fig:cifar_scaling_with_depth}.}
    \label{fig:cifar_hyp_with_depth}
\end{figure}

We then use the values obtained for $\alpha$ and $\beta$ to verify our hypotheses. Figure~\ref{fig:cifar_hyp_with_depth} shows that both the $\alpha$-scaled norm of increments of $\Delta^{(L)}$ and the $\beta$-scaled norm of increments of $A^{(L)}$ seem to have lower bounds as the depth grows. This suggests that Hypothesis~\ref{hypothesis.1} does not hold for convolutional layers.

We also observe that the root sum of squares stays in the same order as the depth increases. Coupled with the fact that the maximum norms of $\Delta^{(L)}$ and $A^{(L)}$ are close to constant order as the depth increases, this suggests that the scaling limit is sparse with a finite number of weights being of constant order in $L$.

\subsection{Summary: three scaling regimes \label{sec:exp_summary}}

Our experiments show different scaling behaviors depending on the network architecture, especially the smoothness of the activation function. In Section~\ref{sec:fully_connected}, for fully-connected layers with $\tanh$ activation and a common pre-factor $\delta^{(L)}$ across layers, we observe a behavior consistent with Hypothesis~\ref{hypothesis.1} for both the synthetic dataset and MNIST. 
In contrast, a ResNet with fully-connected layers with $\relu$ activation and $\delta_k^{(L)}\in\R$ shows behavior compatible with Hypothesis~\ref{hypothesis.2} both for the synthetic dataset and MNIST. 

In the case of convolutional architectures trained on CIFAR-10 (Section~\ref{sec:classification}) we observe that the maximum norm of the trained weights does not decrease with the network depth and the trained weights display a sparse structure, indicating a third scaling regime corresponding to a sparse structure for both $\Delta^{(L)}$ and $A^{(L)}$. These results are consistent with previous evidence on the existence of sparse CNN representations for image recognition~\cite{mallat2016}. We stress that the setup for our CIFAR-10 experiments has been chosen to approach state-of-the-art test performance with our generic architecture, as shown in Appendix~\ref{sec:cifar_classification_results}.


Note that the reason we consider networks  with many layers (up to $L=10321$) is to investigate the behavior of coefficients as depth varies, not because of any claim that very deep networks are more robust or generalize better than shallower ones. In fact, most of the networks exhibit good accuracy for  depths $L \geq 15$.

%% file: sec/4_asymptotics.tex
\section{Deep Network Limit} \label{sec:results}
In this section, we analyze the scaling limit of the residual network \eqref{forward-map-resnet} under  Hypotheses \ref{hypothesis.1} and  \ref{hypothesis.2}.

\subsection{Setup and assumptions} \label{sec:setup-assumptions}
We consider $\delta^{(L)} =L^{-\alpha}$ for some $\alpha \ge 0$ and
\begin{equation} \label{eq:resnet.v2}
\begin{aligned}
h_0^{(L)} &= x, \\
h^{(L)}_{k+1} &= h^{(L)}_{k} + L^{-\alpha}
\,\sigma_d\left(A^{(L)}_k h^{(L)}_{k}+ b^{(L)}_k\right),
\end{aligned}
\end{equation}
with
\begin{eqnarray*}
A^{(L)}_k &=& L^{-\beta}\overline{A}_{k/L} + W^A_{(k+1)/L}-W^A_{k/L}, \\
b^{(L)}_k &=& L^{-\beta}\overline{b}_{k/L} + W^b_{(k+1)/L}-W^b_{k/L},
\end{eqnarray*}
where $(W^A_t)_{t\in [0,1]}$ and $(W^b_t)_{t\in[0,1]}$ are Itô processes \cite{revuz2013continuous} with regularity conditions specified in Appendix \ref{app:setup}. 

\begin{remark}
Hypothesis \ref{hypothesis.1} corresponds to the case $W^A \equiv 0$ and $W^b \equiv 0$. Hypothesis \ref{hypothesis.2} corresponds to the case where $W^A$ and $W^b$ are non-zero. 
\end{remark}
We use the following notation for the quadratic variation of $W^A$ and $W^b$:
\begin{align}
\left[W^A\right]_t=\int_0^t \Sigma^A_u \dd u,\quad \left[W^b\right]_t=\int_0^t \Sigma^b_u \dd u, \label{eq.qv}
\end{align}
where $\Sigma^A$ and $\Sigma^b$ are bounded processes with values respectively in $\mathbb{R}^{d,\,\otimes 4}$ and $\mathbb{R}^{d \times d}$. Let $Q_i:[0,1]\times\mathbb{R}^{d} \rightarrow \mathbb{R}$ be the (random) quadratic form defined by
\begin{equation} \label{eq:Q_thm}
    Q_i(t,x)\coloneqq \sum_{j,k=1}^d x_j x_k \left(\Sigma^A_t\right)_ {ijik} + \Sigma^b_{t,ii}.
\end{equation}
and $Q(t,x)=(Q_1(t,x), \ldots, Q_d(t,x))$. Our analysis focuses on smooth activation functions.



\begin{assumption}[Activation function]\label{ass:activation}
The activation function $\sigma$ is in $\CC^3(\mathbb{R},\mathbb{R})$ and satisfies $\sigma(0)=0$, $\sigma^{\prime}(0)=1$. Moreover, $\sigma$ has a bounded third derivative  $\sigma^{\prime\prime\prime}$.
\end{assumption}
Most smooth activation functions, including $\tanh$, satisfy this condition. 
Also, the boundedness of the third derivative $\sigma^{\prime\prime\prime}$ could be further relaxed to some exponential growth condition, see \cite{peluchetti2020}.


Finally, we assume that the hidden state dynamics $({h}^{(L)}_k, k=1, \ldots, L)$ given by \eqref{forward-map-resnet} is uniformly integrable. (For a precise statement see Assumption \ref{ass:strong} in Appendix.) This is a reasonable assumption since both the inputs and the outputs of the network are uniformly bounded.

\subsection{Informal derivation of the deep network limit}\label{sec:informal_derivation}


We first provide an informal analysis on the derivation of the deep network limit. Denote $t_k=k/L$ and define for $s\in [t_k,t_{k+1}]$:
\begin{align*}
M^{(L),k}_s &= \left(W^A_s- W^A_{t_k}\right) h^{(L)}_{k} + W^b_s- W^b_{t_k} \\
&+L^{1-\beta} \overbar{A}_{t_k} h^{(L)}_{k}(s-t_k)+L^{1-\beta} \overline{b}_{t_k}(s-t_k).
\end{align*}
When $\beta=0$, we need $\alpha=1$ to obtain a non-trivial limit. In this case, the noise terms in $M^{(L), k}_s$ are vanishing as $L$ increases, and the limit is the Neural ODE. See Lemma 4.6 in \cite{TvG2018}.

When $\beta > 0$ we can apply the Itô formula~\cite{ito1944} to $\sigma \big( M^{(L),k}_s\big)$ for $s\in [t_k,t_{k+1})$ to obtain the approximation
\begin{equation} \label{eq:euler_maruyama3} 
    h^{(L)}_{k+1}-h^{(L)}_k = \delta^{(L)} \sigma \left( M_{t_{k+1}}^{(L),k}\right)  = D_1 + D_2 + D_3 + o\left(1/L\right)
\end{equation}
where
\begin{align*}
D_1 &= L^{-\alpha}\left(\left(W^A_{t_{k+1}}-W^A_{t_{k}}\right)h_k^{(L)}+\left(W^b_{t_{k+1}}-W^b_{t_{k}}\right)\right) \\
D_2 &= \frac{1}{2} L^{-\alpha} \sigma^{\prime \prime}(0)\,Q\left(t_k,{h^{(L)}_k}\right)({t_{k+1}-t_{k}})\\
D_3 &= L^{1-\beta-\alpha}\left(\overbar{A}_{t_k}h_{k}^{(L)}({t_{k+1}-t_{k}})+\overline{b}_{t_k}({t_{k+1}-t_{k}})\right).
\end{align*}
When $\alpha = 0$, we see from $D_1$ that~\eqref{eq:euler_maruyama3} admits a diffusive limit, that is with non-vanishing noise terms $W^A$ and $W^b$. In this case, $D_2$ stays bounded when $L$ increases, and $D_3$ does not explode only when $\beta \geq 1$. The case $\alpha=0, \beta \geq 1$ corresponds to a stochastic differential equation (SDE) limit.
 
When $\alpha > 0$, $D_1$ and $D_2$ vanish in the limit $L\to\infty$, and we need $\beta = 1-\alpha$ to obtain a non-trivial ODE limit.

We provide precise mathematical statements of these results in the next section.
\subsection{Statement of the results}
The following results describe the different scaling limits of the hidden state dynamics $(h^{(L)}_k, k=1, \ldots, L)$ for various values of scaling exponents $\alpha$ and $\beta$.

First, we show that Hypothesis~\ref{hypothesis.1} with a smooth activation function leads to convergence in sup norm to an ODE limit that is different from the neural ODE behavior described in~\cite{CRBD2018,TvG2018,HR2018}.
\begin{theorem}[Asymptotic behavior under Hypothesis~\ref{hypothesis.1}]\label{thm:H1} If the activation function satisfies Assumption \ref{ass:activation},
 $0<\alpha < 1$ and $\alpha+\beta=1$, then the   hidden state  converges 
to the solution to the ODE
   \begin{eqnarray}\label{eq:limit5}
  \frac{\dd H_t}{\dd t} = \overbar{A}_t H_t+\overline{b}_t,\qquad H_0=x,
   \end{eqnarray}
 in the sense that  $$\mathop{\lim}_{L \rightarrow \infty} \mathop{\sup}_{0 \leq t \leq 1}\left|\left|H_t-h^{(L)}_{\floor{t L}}\right|\right|=0.$$
\end{theorem}
In particular, this implies the convergence of the hidden state process 
for any typical initialization, i.e almost-surely with respect to the initialization.

Note that in Theorem~\ref{thm:H1}, the limit~\eqref{eq:limit5} defines a linear input-output map behaving like a linear network~\cite{ACGH2019}. This is different from the neural ODE~\eqref{eq:neural_ode_limit}, where the activation function $\sigma$ appears in the limit. Interestingly, the limit is a controlled ODE where the control parameters are linear in the derivative of the state. Their expressivity is discussed in~\cite{CLT2019}.



We show that under Hypothesis~\ref{hypothesis.2}, we may obtain either an SDE or an ODE limit. In the latter case, the limiting ODE is found to be different from the neural ODE~\eqref{eq:neural_ode_limit}.
\begin{theorem}[Asymptotic behavior under Hypothesis~\ref{hypothesis.2}]\label{thm:H2} 
Let $\sigma$ be an activation function satisfying Assumption \ref{ass:activation}.
\underline{\emph{$\alpha=0$ and $\beta \ge 1$: diffusive limit}}. Let $H$ be the solution to the SDE
\begin{equation} \label{eq:limit1}
   \begin{aligned}
   \dd H_t &= H_t\,\dd W^A_t +\dd W^b_t + \frac{1}{2}\sigma^{\prime \prime}(0)Q(t,H_t)\,\dd t \\ &+\one_{\beta=1}(\overbar{A}_t H_t+\overline{b}_t)\,\dd t,
   \end{aligned}
\end{equation}
with initial condition $H_0=x$.
If   there exists $p_2>2$  such that $\mathbb{E}\left[\sup_{0 \leq t \leq 1}\|H_t\|^{p_2} \right]  <\infty $, then the   hidden state   converges uniformly in $L^2$ to the solution $H$ of~\eqref{eq:limit1}:
  $$ \lim_{L \to \infty}\mathbb{E}\left[\sup_{0\leq t\leq 1}\norm{h^{(L)}_{\floor{t L}} - {H}_t}^2\right] = 0.$$


\underline{$0<\alpha<1$, $\alpha + \beta = 1$: \emph{ODE limit}}. The   hidden state    converges uniformly in $L^2$ to the solution of the ODE
\begin{eqnarray}\label{eq:limit3}
  \frac{\dd H_t}{\dd t} = \overbar{A}_t H_t+\overline{b}_t,
   \end{eqnarray}
with initial condition $H_0=x$:
 $$ \lim_{L \to \infty}\mathbb{E}\left[\sup_{0\leq t\leq 1}\norm{h^{(L)}_{\floor{t L}} - {H}_t}^2\right] = 0.$$
\end{theorem}
Note that we prove uniform convergence in $L^2$, also known as strong convergence.

The detailed assumptions and a sketch of the proof for Theorems~\ref{thm:H1} and~\ref{thm:H2} are given in Appendix~\ref{app:proof}. Further details and some extensions may be found in the companion paper~\cite{CCRX2021}.

The idea of the proof is to view the ResNet as a {\it nonlinear} Euler discretization of the limit equation, and then bound the difference between the hidden state and a classical Euler discretization.
Then, using an extension of the techniques in \cite{higham2002strong} to the case of equations driven by It\^o processes, we show strong convergence in the following way.  
We first show that the drift term of~\eqref{eq:limit1} is locally Lipschitz (Appendix~\ref{app:theorem2}). We then prove the strong convergence of the hidden state dynamics \eqref{eq:resnet.v2} by bounding the difference between the hidden state and an Euler scheme for the  limiting equation. It is worth mentioning that the convergence results in \cite{higham2002strong} hold for a class of time-homogeneous diffusion processes whereas our result holds for general It\^o processes. This distinction is important for training neural networks since the diffusion assumption involves the Markov property which cannot be expected to hold after training with backpropagation.

In addition, we also relax a technical condition from~\cite{higham2002strong}, which is difficult to verify in practice. See Remark \ref{rmk:uniform_integrability}.

\subsection{Remarks on the results}

Interestingly, when the activation function $\sigma$ is smooth, all limits in both Theorems~\ref{thm:H1} and~\ref{thm:H2} depend on the activation only through $\sigma^{\prime}(0)$ (assumed to be $1$ for simplicity) and $\sigma^{\prime\prime}(0)$. Hypotheses~\ref{hypothesis.1} and~\ref{hypothesis.2} lead to the same ODE limit when $0<\alpha<1$ and $\alpha+\beta=1$. In contrast to  the neural ODE~\eqref{eq:neural_ode_limit}, the characteristics of $\sigma$ away from $0$ are not relevant to this limit.  
In addition, our proof relies on the smoothness of $\sigma$ at $0$. If the activation function is not differentiable at $0$, then a different limit should be expected.

The case $\overbar{A}\equiv 0$, $\overline{b}\equiv 0$, $\alpha=0$, and $\beta=1$ in Theorem~\ref{thm:H2} is considered in~\cite{peluchetti2020}, who prove weak convergence under the additional assumption that $W^A$ and $W^b$ are Brownian motions with constant drift. In our setup, $W^A$ and $W^b$ are allowed to be arbitrary Itô processes, whose increments, i.e. the network weights, are not necessarily independent nor identically distributed. This allows for a general distribution and dependence structure of weights across layers.

The concept of weak convergence used in~\cite{peluchetti2020}  corresponds to convergence of quantities averaged across many random IID weight initializations. In practice, as the training is done only once, the strong convergence, shown in Theorems~\ref{thm:H1} and~\ref{thm:H2} is a more relevant notion for studying the asymptotic behavior of deep neural networks.

\subsection{Link with numerical experiments}

Let us now discuss how the analysis above sheds light on the numerical results in Section \ref{sec:fully_connected} and Section~\ref{sec:classification}. 


Figure~\ref{fig:scaling_tanh_shared_delta} shows that $\beta \approx 0.3$ and $\alpha \approx 0.7$ for the synthetic dataset with fully-connected layers and $\tanh$ activation function, and Figure~\ref{fig:hypothesis_tanh_shared_delta} suggests that Hypothesis~\ref{hypothesis.1} is more likely to hold. This corresponds to the assumptions of Theorem~\ref{thm:H1} with the ODE limit~\eqref{eq:limit5}. This is also consistent with the estimated decomposition in Figure~\ref{fig:hypothesis_tanh_shared_delta} (right) where the noise part tends to be negligible. 

In the case of $\relu$ activation with fully-connected layers, we observe that $\beta+\alpha \approx 1$ from Figure~\ref{fig:scaling_relu_scalar_delta} (left). Since ReLU is homogeneous of degree 1 (see Remark~\ref{rmk:relu}), $|\delta^{(L)}|$ can be moved inside the activation function, so without loss of generality we can assume $\alpha=0$ and $\beta \approx 1$. 
If we replace the $\relu$ function by a smooth version $\sigma^\epsilon$, then the limit is described by the stochastic differential equation~\eqref{eq:limit1}.
The $\relu$ case would then correspond to a limit of this equation as $\epsilon \to 0$. The existence of such a limit is, however, nontrivial. 

From the experiments with convolutional architectures, we observe that the maximum norm  (Figure~\ref{fig:cifar_scaling_with_depth}), the scaled norm of the increments, and the root sum of squares (Figure~\ref{fig:cifar_hyp_with_depth}) are upper bounded as the number of layers $L$ increases. This indicates that the weights become sparse  when $L$ is large. In this case, there is no continuous ODE or SDE limit and Hypotheses~\ref{hypothesis.1} and~\ref{hypothesis.2} both fail. 
This emergence of sparse representations in convolutional networks is consistent with previous results on such networks \cite{mallat2016}.

%% file: sec/5_conclusion.tex
\section{Conclusion} \label{sec:conclusion}
We study the scaling behavior of trained weights in deep residual networks. We provide evidence for the existence of at least three different scaling regimes that encompass differential equations and sparse scaling limits. We also theoretically characterize the ODE and SDE limits for the hidden state dynamics in deep fully-connected residual networks.

Our work contributes to a better understanding of the behavior of residual networks and the role of network depth. Our findings point to interesting questions regarding the asymptotic behavior of such networks in the case of non-smooth activation functions and more complex architectures.


%% file: sec/appendix-final.tex
\appendix

\section{Hyperparameters \label{sec:hyperparameters}}

We provide in Table~\ref{tab:hyperparameters} the training hyperparameters used in our numerical experiments. In Table~\ref{tab:description_hyperparameters}, we give a short description of each hyperparameter. For the convolutional architecture, we also use a momentum of 0.9, a weight decay of $0.0005$ and a cosine annealing learning rate scheduler~\cite{DBLP:journals/corr/LoshchilovH16a}.

{
    \vspace{0.2cm}
    \tabulinesep=1.3mm
        \captionof{table}{Training hyperparameters.}
    \begin{center}
    \begin{tabu}{c|c|c|c|c|c|c|c|c|c}
        \hline
        Dataset & Layers & $N$ & $B$ & $\lr$ & $L_{\min}$ & $L_{\max}$ & $T_{\max}$ & $N_\text{epochs}$ & $\epsilon$ \\ \hline \hline
        Synthetic & Fully-connected & 1,024 & 32 & 0.01 & 3 & 10,321 & 160 & 5 & 0.01 \\ \hline
        MNIST & Fully-connected & 60,000 & 50 & 0.01 & 3 & 942 & 12,000 & 10 & 0.01  \\ \hline
        CIFAR-10 & Convolutional & 60,000 & 128 & 0.1 & 8 & 121 & 93,800 & 200 & None \\ \hline
    \end{tabu}
    \end{center}
    \label{tab:hyperparameters}

    \vspace{0.2cm}
}

{
    \vspace{0.2cm}
    \tabulinesep=1.3mm
        \captionof{table}{Description of the values in Table~\ref{tab:hyperparameters}. Note that $T_{\max} = \ceil{ \frac{N}{B} } N_\text{epochs}$.}
    \begin{center}
        \begin{tabu}{c|l}
            \hline
            Parameter & Description \\  \hline
            $N$ & number of training samples \\
            $B$ & minibatch size \\
            $\lr$ & learning rate \\
            $L_{\min}$ & smallest network depth \\
            $L_{\max}$ & largest network depth \\
            $T_{\max}$ & max number of SGD updates \\
            $N_\text{epochs}$ & max number of epochs \\
            $\epsilon$ & early stopping value \\
            \hline
        \end{tabu}
    \end{center}

    \label{tab:description_hyperparameters}

    \vspace{0.8cm}
}

We report in Table \ref{hyperparam-sensi} below results for $\tanh$ and trainable $\delta$  on the synthetic data with different batch sizes and learning rates ($5$ different seeds). 
We observe that the learning rate does affect $\alpha$ and $\beta$, but their sum always stays around $1$. The batch size has no effect on the exponents.
    \captionof{table}{Average value of $\alpha$ (left) and $\beta$ (right) for the trained weights, over $5$ random initializations, $\eta$ is the learning rate, and $B$ the batch size.}
    \label{hyperparam-sensi}
    \begin{minipage}[t]{0.48\textwidth}
        \centering
        \tabulinesep=1.5mm
        \setlength\tabcolsep{6pt}
        \begin{tabu}{c|c|c|c}
            \hline
            $\alpha$ & $B=8$ & $B=32$ & $B=128$  \\ \hline
            $\eta=.01$ & $.69 \pm .02$ & $.73\pm .02$ & $.67 \pm .02$  \\ \hline
            $\eta=.003$ & $.59\pm .05$ & $.60\pm .01$ & $.58\pm .01$ \\ \hline
            $\eta=.001$ & $.58\pm .01$ & $.55\pm .01$ & $.53 \pm .01$ \\ \hline
        \end{tabu}
    \end{minipage}
    \hfill 
    \begin{minipage}[t]{0.48\textwidth}
        \centering
        \tabulinesep=1.4mm
        \setlength\tabcolsep{6pt}
        \begin{tabu}{c|c|c|c}
            \hline
            $\beta$ & $B=8$ & $B=32$ & $B=128$  \\ \hline
            $\eta=.01$ & $.24\pm .02$ & $.29 \pm .05$ & $.22\pm .02$    \\ \hline
            $\eta=.003$ & $.33\pm .01$ & $.41\pm .06$ & $.40 \pm .02$ \\ \hline
            $\eta=.001$ & $.39\pm .02$ & $.43\pm .02$ & $.41 \pm .01$  \\ \hline
        \end{tabu}
    \end{minipage}
\vspace{0.5cm}

\section{Scaling Analysis of the Biases $b^{(L)}$ in the Fully-Connected Case} \label{app:further_plots}

\begin{figure}[h!]
    \centering
    \includegraphics[width=0.33\textwidth]{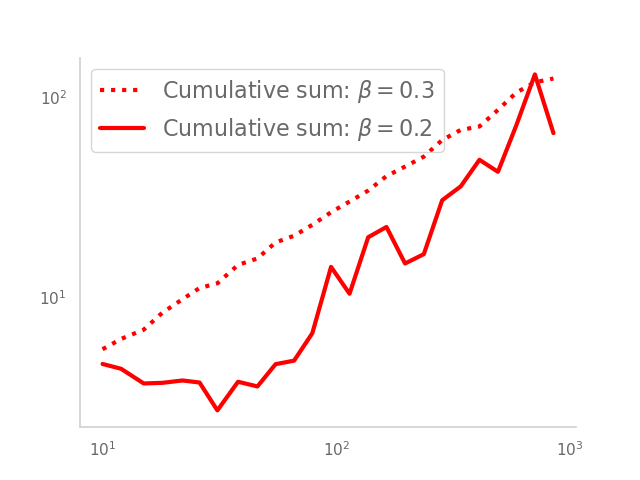}
    \includegraphics[width=0.33\textwidth]{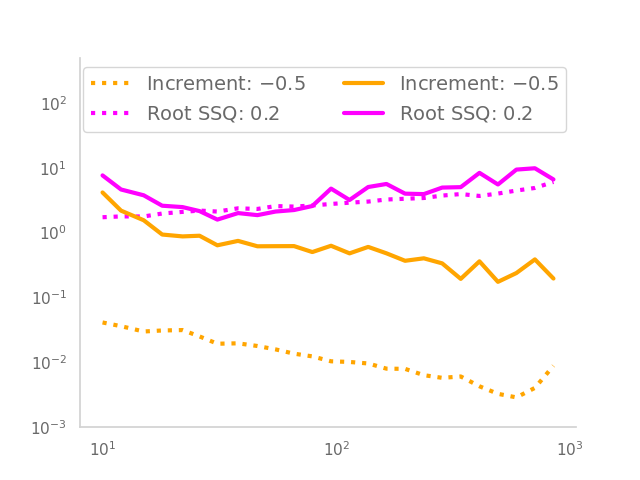}
    \includegraphics[width=0.33\textwidth]{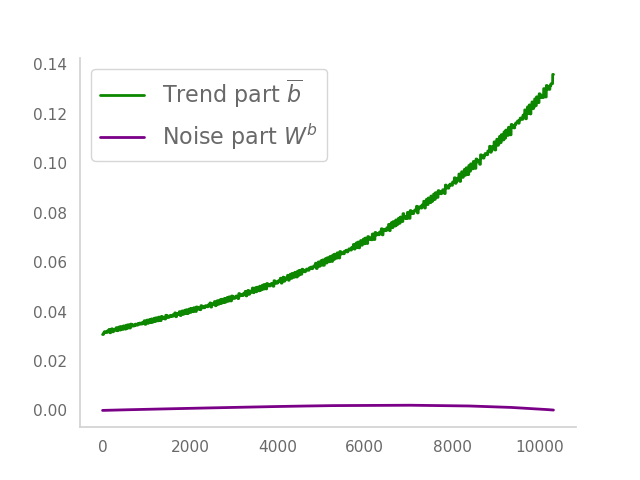}
    \caption{Scaling and hypothesis verification for $\tanh$ activation and $\delta^{(L)}\in\R$. Left: Maximum norm of $b^{(L)}$ with respect to $L$, in log-log scale. Middle: we plot in log-log scale the root sum of squares of $b^{(L)}$ in pink and the $\beta-$scaled norm of increments of $b^{(L)}$ in orange. The dashed lines are for the synthetic data and the solid lines are for MNIST. Right: Decomposition of the trained weights $b^{(L)}_{k, 5}$ with the trend part $\overbar{b}$ and the noise part $W^b$ for $L=10321$, as defined in~(\ref{scaling-W}), for the synthetic dataset.}
    \label{fig:analysis_b_tanh}
\end{figure}

\begin{figure}[h!]
    \centering
    \includegraphics[width=0.33\textwidth]{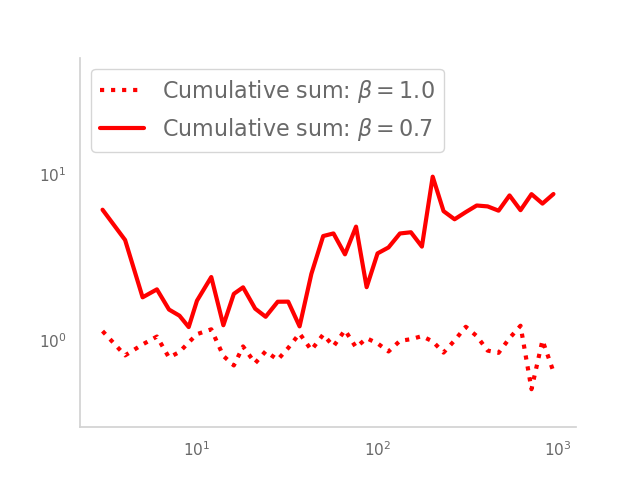}
    \includegraphics[width=0.33\textwidth]{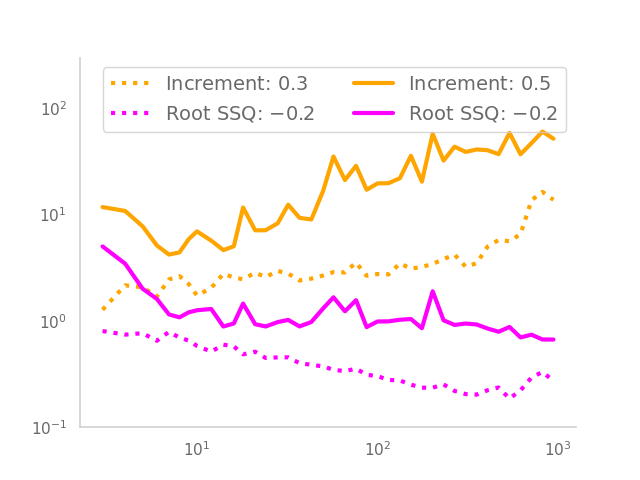}
    \includegraphics[width=0.33\textwidth]{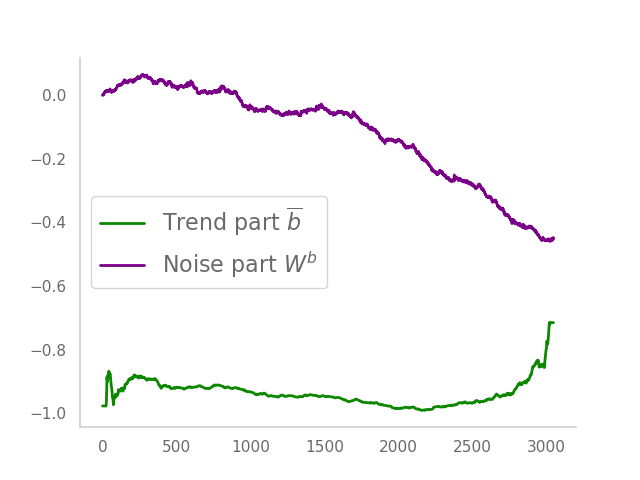}
    \caption{Scaling and hypothesis verification for $\relu$ activation and $\delta_k^{(L)}\in\R$. Left: Maximum norm of $\vert \delta^{(L)}\vert b^{(L)}$ with respect to $L$, in log-log scale. Middle: we plot in log-log scale the root sum of squares of $\vert \delta^{(L)}\vert b^{(L)}$ in pink and the $\beta-$scaled norm of increments of $\vert \delta^{(L)}\vert b^{(L)}$ in orange. The dashed lines are for the synthetic data and the solid lines for MNIST. Right: Decomposition of the trained weights $ \vert \delta^{(L)} \vert \, b^{(L)}_{k, 6}$ with the trend part $\overbar{b}$ and the noise part $W^b$ for $L=10321$, as defined in~(\ref{scaling-W}), for the synthetic dataset.}
    \label{fig:analysis_b_relu}
\end{figure}

We mention in the main text that the behaviour of the trained values of $b^{(L)}$ with the depth $L$ is similar to that of $A^{(L)}$, in the fully-connected case of Section~\ref{sec:fully_connected}. We verify these claims here. To do so, we follow the same methodology outlined in Section~\ref{sec:procedure-experiments} for $b^{(L)}$, and we show the results for the $\tanh$ case in Figure~\ref{fig:analysis_b_tanh} and for the $\relu$ case in Figure~\ref{fig:analysis_b_relu}. We observe that the maximum norm, the scaled norm of the increments and the root sum of squares of $b^{(L)}$ scales in the same way as $A^{(L)}$ as the depth $L$ increases. In particular, the scaling exponent $\beta$ for $b^{(L)}$ is equal to the scaling exponent of $A^{(L)}$, justifying the setup considered in Section \ref{sec:setup-assumptions}.  

\section{Convolutional Network Results on CIFAR-10 \label{sec:cifar_classification_results}}

We give in Table~\ref{tab:cifar_test_acc} the final test accuracy of our convolutional residual networks trained on an NVIDIA GeForce RTX 2080 GPU. The results are in line with those of traditional ResNet architectures~\cite{HZRS2016}, even though our networks do not have batch normalization layers~\cite{DBLP:journals/corr/IoffeS15}. It is also noteworthy to add that our concept of depth is not that of traditional ResNets. We define the number of layers $L$ as the number of skip connections in the network, that is the number of $\Delta_k$ kernels in~\eqref{eq:2d-forward-map}.

We also note that the test error in Table~\ref{tab:cifar_test_acc} does not decrease with network depth. This is due to the fact, already mentioned in Section~\ref{sec:exp_summary}, that smaller depths usually suffice to get a good accuracy. In our case, we focus on a simple setting that still approaches the results obtained in practice. Rather than trying to find a setup that maximizes the accuracy, for instance with batch normalization or the Adam optimizer~\cite{kingma2014method}, we aim to understand the scaling of residual networks in practical cases.

{
    \tabulinesep=1.3mm
        \captionof{table}{Test error in $\%$ on CIFAR-10 for each network depth $L$.}
    \begin{center}
    \begin{tabu}{c|c|c|c|c|c|c|c|c}
        \hline
        $L$ & 8 & 11 & 12 & 14 & 16 & 20 & 24 & 28 \\ 
        \hline
        Test error & 6.64 & 6.37 & 6.32 & 5.98 & 6.25 & 5.98 & 6.24 & 7.03 \\ 
        \hline \hline
        $L$ & 33 & 42 & 50 & 65 & 80 & 100 & 121 & \\ 
        \hline
        Test error & 6.13 & 6.21 & 6.32 & 6.19 & 6.30 & 6.20 & 6.37 & \\
        \hline
    \end{tabu}
    \end{center}
    \label{tab:cifar_test_acc}

    \vspace{0.2cm}
}

\section{Residual Network Architecture \label{sec:resnet_architectures}}

\begin{figure}[h!]
    \centering
    \includegraphics[width=1.0\textwidth]{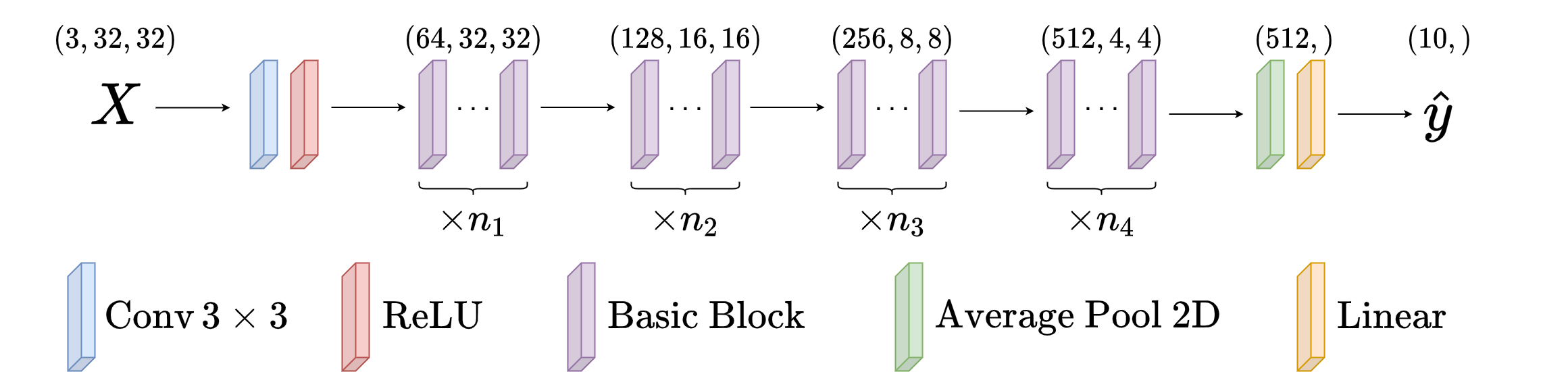}
    \vspace{0.1in}
    \caption{Residual architecture. There are 4 blocks that are respectively repeated $n_1$, $n_2$, $n_3$ and $n_4$ times. The network depth is $L=n_1+n_2+n_3+n_4$. The Basic Block architecture is detailed in Figure~\ref{fig:cifar_net_block}.}
    \label{fig:cifar_net_basic}
\end{figure}

\begin{figure}[h!]
    \centering
    \includegraphics[width=1.0\textwidth]{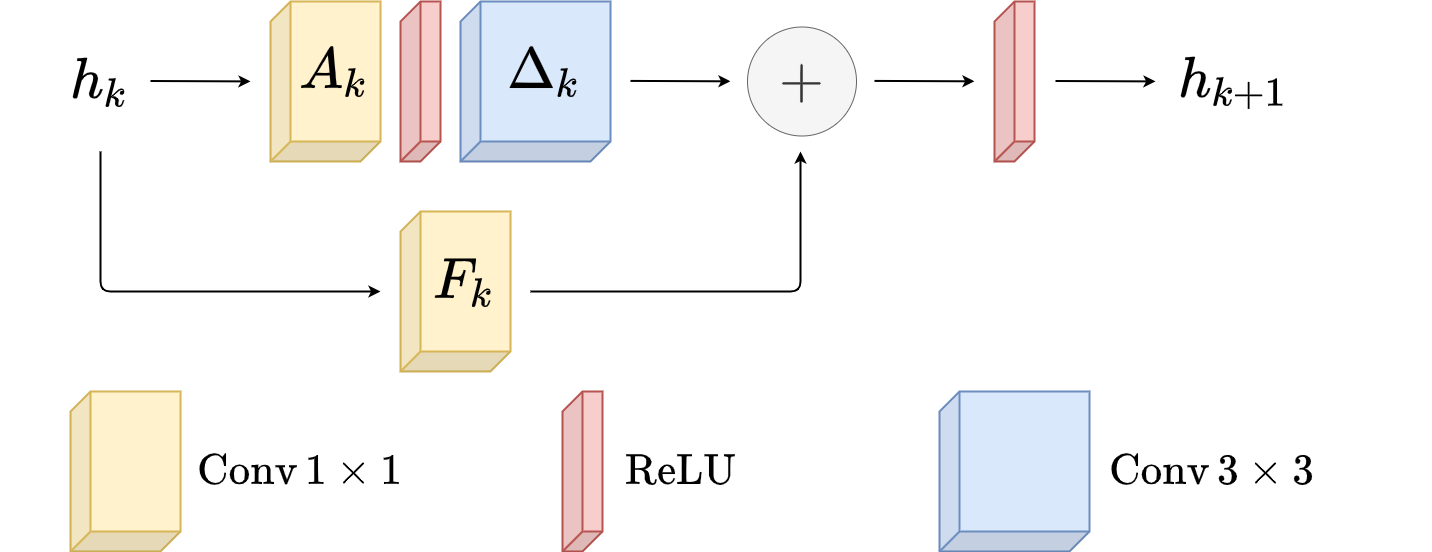}
    \vspace{0.1in}
    \caption{Basic Block from Figure~\ref{fig:cifar_net_basic}. See~\eqref{eq:2d-forward-map} for details.}
    \label{fig:cifar_net_block}
\end{figure}

\newpage
\section{Proofs of Technical Results in Section \ref{sec:results}}\label{app:proof}
This appendix outlines the main arguments in the proofs of results stated in Section \ref{sec:results}. Further mathematical details are provided in \cite{CCRX2021}.

\subsection{Setup}\label{app:setup}
As specified in Section \ref{sec:setup-assumptions}, we model the cumulative sum of weights (resp. bias) as It\^o processes $(W^A_t)_{t \ge 0}$ (resp. $(W^b_t)_{t \ge 0}$)  on some filtered probability space
$(\Omega, \mathcal{F}, \mathbb{F}=(\mathcal{F}_t)_{t\geq 0}, \mathbb{P})$. 
This means   $W^A,W^b$ satisfy \begin{equation}
\begin{aligned}
\left(\dd W_t^A\right)_{ij} &= \left(U^A_t\right)_{ij} \dd t + \sum_{k,l=1}^d \left(q_t^A\right)_{ijkl} \left(\dd B^{A}_t\right)_{kl} \quad \mbox{for }\, i,j=1, \ldots, d ,\\
\dd W_t^b &= U^b_t \dd t +q^b_t \dd B^b_t,
\end{aligned}
\end{equation} where
$(B^A_t)_{t\ge 0}$, resp. $(B^b_t)_{t\ge 0}$ are $d\times d$-dimensional, resp. $d$-dimensional,  Brownian motions,
 and $q_t^A\in \mathbb{R}^{d,\otimes4}$ and $q_t^b\in \mathbb{R}^{d\times d}$ for $t\in [0,1]$. We set  $W_0^A = 0$, $W_0^b = 0$.
 Denote the quadratic variation processes as:
\begin{equation}
\begin{aligned} \label{def:covariance-tensors}
\left(\Sigma_t^A\right)_{i_1 j_1 i_2 j_2} &\coloneqq \sum_{k,l=1}^d \left(q_t^A\right)_{i_1 j_1 kl} \, \left(q_t^A\right)_{i_2 j_2 kl}, \quad \mbox{for } \, i_1, j_1, i_2, j_2 = 1, \ldots, d,\\
\Sigma_t^b &\coloneqq q_t^b \left(q_t^b\right)^{\top}.
\end{aligned}
\end{equation}
We assume 
$(U_t^A)_{t\geq 0}$, $(U_t^b)_{t\geq 0}$, $(\Sigma_t^A)_{t\geq 0}$ and $(\Sigma_t^b)_{t\geq 0}$ are progressively measurable processes satisfying:
\begin{assumption}[Regularity assumptions] \label{ass:ito} We assume 
\begin{itemize}
    \item[(i)] There exists a constant $C_1>0$ such that almost surely
\begin{eqnarray}\label{eq:bound_ito}
\sup_{0 \leq t \leq 1}\norm{U_t^A} + \sup_{0 \leq t \leq 1} \norm{U_t^b} + \sup_{0 \leq t \leq 1} \norm{\Sigma_t^A} + \sup_{0 \leq t \leq 1} \norm{\Sigma_t^b} \leq C_1.
\end{eqnarray}
\item[(ii)]  There exist $M>0$ and $\kappa>0$ such that $\forall s,t\in[0,1]$ almost surely
 \begin{eqnarray}\label{eq:continuity_ito}
\norm{U_t^A-U^A_s}^2 + \norm{U_t^b-U^b_s}^2+ \norm{\Sigma_t^A-\Sigma^A_s}^2 +\norm{\Sigma_t^b-\Sigma^b_s}^2  \leq M |t-s|^{\kappa}\\
\label{eq:continuity_func}
\norm{\bar{A}_t-\bar{A}_s}^2 +\norm{\bar{b}_t-\bar{b}_s}^2 \leq M |t-s|^{\kappa}.
  \end{eqnarray}
\end{itemize}
\end{assumption}

\begin{lemma}[Uniform integrability]\label{lemma:uni_integrability_W}
Under Assumption \ref{ass:ito}-(i), we have, for any $p_0> 1$
\begin{equation}\label{eq:uniform_W}
\mathbb{E}\left[\sup_{0\leq s \leq 1}\norm{W^A_s}^{p_0}\right], \,\,\, \mathbb{E}\left[\sup_{0\leq s \leq 1}\norm{W^b_s}^{p_0}\right]<\infty.
\end{equation}
\end{lemma}
Lemma \ref{lemma:uni_integrability_W} is proven by first applying Minkowski inequality  to $\mathbb{E}\left[\sup_{0\leq s \leq 1}\norm{W^A_s}^{p_0}\right]$ and then applying Burkholder-Davis-Gundy inequality to  $\mathbb{E}\left[\sup_{0\leq s \leq 1}\norm{\left(\int_0^s \sum_{k,l=1}^d \left(q_t^A\right)_{ijkl} \left(\dd B^{A}_t\right)_{kl}\right)_{i,j}}^{p_0} \right]$.

\begin{assumption}[Uniform integrability]\label{ass:strong}
There exist $p_1 >4$ and a constant $C_0$ such that for all $L$,
 \begin{eqnarray}\label{ass:boundedness}
 \mathbb{E}\left[\sup_{0 \leq k \leq L}\norm{h^{(L)}_k}^{p_1} \right]  \leq C_0.
 \end{eqnarray}
 \end{assumption}

\subsection{Proof of Theorem \ref{thm:H1}}
We now provide a sketch of the proof for Theorem \ref{thm:H1} under Assumption \ref{ass:activation}, Assumption \ref{ass:ito} with $W^A\equiv 0$ and $W^B \equiv 0$, and Assumption \ref{ass:strong}. 
The detailed proof can be found in a companion paper \cite{CCRX2021}.
Under Assumption \ref{ass:strong}, there exists $C_{\infty}>0$ such that $\sup_{L\in\N} \max_{k=1,2,\ldots,L}\norm{h_k^{(L)}} \leq C_{\infty}.$
Denoting $\Delta h_{k}^{L}:= h_{k+1}^{L}-h_{k}^{L}$ and $M^{(L)}_k(h) := \overbar{A}_{t_k}h+\overbar{b}_{t_k}$,
 from~\eqref{eq:euler_maruyama3} we have
$$\Delta h_k^{(L)} := h_{k+1}^{(L)}-h_k^{(L)} = L^{-\alpha}\sigma \left(  L^{-\beta}M_k^{(L)}(h_k^{(L)})\right).$$
For any vector $x\in \mathbb{R}^d$, denote $(x)_i$ as the $i$-th component of $x$. Further denote $\Delta h_k^{(L),i}$ and $M_{k}^{(L),i}$ the $i$-th element of $\Delta h_k^{(L)}$ and $M_{k}^{(L)}$, respectively. 
Applying a third-order Taylor expansion of $\sigma$ around $0$ using Assumption~\ref{ass:activation} we get
\begin{eqnarray}
\Delta h_{k}^{L,i} =  L^{-1} M_k^{(L),i}(h_k^{(L)})+ \frac{1}{2}\sigma^{\prime \prime}(0)L^{-\beta-1} \left(M_k^{(L),i}(h_k^{(L)}) \right)^2  + \frac{1}{6}\sigma^{\prime\prime\prime}(\nu_k^i) L^{-2\beta-1}\left(M_k^{(L),i}(h_k^{(L)}) \right)^3\label{eq:delta_h}
\end{eqnarray}
with $|\nu_k^i|\leq L^{-\beta}\left|\left(\overbar{A}_{t_k}h_k^{(L)}+\overbar{b}_{t_k}\right)_i\right|$ under the condition $\alpha+\beta=1$. 
Denote $\{t_k=k/L,\,k=0,1,\ldots,L\}$ as the uniform partition of the interval $[0,1]$. For $t\in [t_k,t_{k+1}]$, define $$\widetilde{H}_t^{(L)} = h_k^{(L)} +(t-t_k) M_k^{(L),i}(h_k^{(L)})+ \frac{1}{2}\sigma^{\prime \prime}(0)L^{-\beta-1} \left(M_k^{(L),i}(h_k^{(L)}) \right)^2  + \frac{1}{6}\sigma^{\prime\prime\prime}(\nu_k^i) L^{-2\beta-1}\left(M_k^{(L),i}(h_k^{(L)}) \right)^3.$$
Then we have  $\widetilde{H}_{t_k}^{(L)}=h_k^{(L)}$ for all $k=0,1,\cdots,L$.

Recall $H_t$ the solution to the ODE \eqref{eq:limit5}. Denote the differences   $d^{(L),1}_{k}(t) = \widetilde{H}_t^{(L)} - h_{k}^{(L)}$ and $d^{(L),2}_{k}(t) = H_t - \widetilde{H}_t^{(L)}$ for $t\in[t_k,t_{k+1}]$. Similarly denote the errors $e^{(L),1}_{k} = \sup_{t_k \leq t \leq t_{k+1}}\norm{d^{(L), 1}_{k}(t)}$ and $e^{(L), 2}_{k} = \sup_{t_k \leq t \leq t_{k+1}}\norm{d^{(L), 2}_{k}(t)}$. 
The proof reduces to showing   $\sup_{1 \leq k \leq L} e^{(L),1}_{k} \rightarrow 0$ and $\sup_{1 \leq k \leq L} e^{(L),2}_{k} \rightarrow 0$ when $L \rightarrow \infty$.

We first bound $e^{(L),1}_{k}$. Denote $c_0 := \sup_{x\in \mathbb{R}}|\sigma^{\prime\prime\prime}(x)|<\infty$.  By definition and direct calculation,
we have $e^{(L),1}_{k} \leq D_{\infty} L^{-1}\label{eq:e_1}$ with constant $ D_{\infty} := A_{\max} C_{\infty} +b_{\max} +\frac{1}{2}\sigma^{\prime \prime}(0)(A_{\max} C_{\infty} +b_{\max} )^2 +\frac{1}{6}c_0(A_{\max} C_{\infty} +b_{\max} )^2$.
Therefore it holds that $\lim_{L \rightarrow \infty}\sup_{1 \leq k \leq L} e^{(L),1}_{k} = 0.$

We next bound $e^{(L),2}_{k}$.  For $t\in[t_k,t_{k+1}]$,
\begin{eqnarray}
d^{(L),2}_{k+1}(t) &=& d^{(L),2}_{k}(t_{k+1}) -(t-t_{k+1}) M_{k+1}^{(L),i}(h_{k+1}^{(L)})+ \int_{t_{k+1}}^{t}(\overbar{A}_sH_s+\overbar{b}_s) \dd s\nonumber\\
&&- \frac{1}{2}\sigma^{\prime \prime}(0)L^{-\beta-1} \left(M_{k+1}^{(L),i}(h_{k+1}^{(L)}) \right)^2  - \frac{1}{6}\sigma^{\prime\prime\prime}(\nu_{k+1}^i) L^{-2\beta-1}\left(M_{k+1}^{(L),i}(h_{k+1}^{(L)}) \right)^3
 \label{eq:d_k}
\end{eqnarray}

From \eqref{eq:delta_h} and \eqref{eq:d_k} we have
\begin{eqnarray*}
e^{(L),2}_{k+1} &\leq& e^{(L),2}_{k} +\sup_{t_{k+1} \leq t \leq t_{k+2}} \norm{ \int_{t_{k+1}}^{t}\left(\left(\overbar{A}_sH_s+\overbar{b}_s \right)- \left( \overbar{A}_{t_{k+1}}h_{k+1}^{(L)}+\overbar{b}_{t_{k+1}}\right) \right) \dd s}\\
&&+  \frac{1}{2}\left|\sigma^{\prime \prime}(0)\right|L^{-\beta-1} \norm{M_{k+1}^{(L)}(h_{k+1}^{(L)}) }^2  + \frac{1}{6}c_0 L^{-2\beta-1}\norm{M_{k+1}^{(L)}(h_{k+1}^{(L)})}^3.
\end{eqnarray*}
As $\beta > 0$, the last two terms are $o(L^{-1})$. Also, direct calculation yields, for $L$ big enough and $A_{\max}:=\sup_{0 \leq t \leq 1}\norm{\overbar{A}_{t}}<\infty$,
\begin{equation*}
\sup_{t_{k+1} \leq t \leq t_{k+2}} \norm{ \int_{t_{k+1}}^{t}\left(\left(\overbar{A}_sH_s+\overbar{b}_s \right)- \left( \overbar{A}_{t_{k+1}}h_{k+1}^{(L)}+\overbar{b}_{t_{k+1}}\right) \right) \dd s}\leq 2 A_{\max} L^{-1}  e_{k+1}^{(L),2} +o(L^{-1}).
\end{equation*}

Finally, $e_0 = \OO(L^{-1})$, so by Grönwall's lemma, we also have $\sup_{1 \leq k \leq L} e^{(L),2}_{k} = \OO(L^{-1})$.

\qedwhite

\subsection{Proof of Theorem \ref{thm:H2}}\label{app:theorem2}
We provide a sketch of the proof for Theorem \ref{thm:H2} under Assumptions \eqref{ass:activation},  \eqref{ass:ito} and  \eqref{ass:strong} for the case $\alpha=0$ and $\beta=1$. Other cases follow similarly. The detailed proof can be found in a companion paper \cite{CCRX2021}.

When $\alpha=0$ and $\beta=1$, we define the targeted SDE limit for the discrete scheme \eqref{eq:resnet.v2} as follows:
\begin{eqnarray}\label{eq:limit_sde}
\dd H_t = \mu(t,H_t)\dd t + \dd V_t^A \, H_t + \dd V_t^b , \quad 0\leq t \leq 1,\,\, \text{ with }\,\, H_0=x,
\end{eqnarray}
in which
\begin{equation} \label{eq:case1_mean}
\begin{aligned}
\mu\left(t,h\right) = U_t^{A} \, h+\,U_t^{b}+ \overline{A}_{t}  h + \overline{b}_{t}+\frac{1}{2}\sigma^{\prime \prime}(0)Q(t,h),\quad
\dd V_t^A =\sum_{k,l=1}^d \left(q_t^A\right)_{ijkl} \left(\dd B_t^A\right)_{kl},\quad \dd V_t^b = q_t^b \,\dd B_t^b,
\end{aligned}
\end{equation}
with $V_0^A=0$ and $V_0^b=0$. 
Here the quadratic variation process $\frac{1}{2}\sigma^{\prime\prime}(0)Q(t,h)$ is the \textit{Itô correction} term for the drift. 

\paragraph{Euler-Maruyama scheme of the limiting SDE.} Denote $\Delta_L =\frac{1}{L}$,  $\{t_k=k/L,\,k=0,1,\ldots,L\}$ and  $\Delta V^A_k = V^A_{t_{k+1}}-V^A_{t_k}$ and $\Delta V^b_k = V^b_{t_{k+1}}-V^b_{t_k}$.  The Euler-Maruyama discretization  of the SDE \eqref{eq:limit_sde} is defined as:
\begin{eqnarray}\label{eq:euler}
\hhat_{k+1}^{(L)}-\hhat_{k}^{(L)} = \mu \left(t_k,\hhat_{k}^{(L)}\right)\Delta_L +\Delta V_k^A  \, \hhat_{k}^{(L)} + \Delta V_k^b  =\hhat_{k}^{(L)} + f^{(L)}\left(k,\hhat_{k}^{(L)}\right)
\end{eqnarray}
where
\begin{eqnarray}
f^{(L)}(k,h) =  \mu \left(t_k,h\right)\Delta_L +\Delta V_k^A  \, h + \Delta V_k^b.
\end{eqnarray}

Define the {\it continuous-time extension} of the hidden state dynamics
\begin{eqnarray}\label{eq:cte}\hbarr^{(L)}_t = h_k^{(L)} {\bf 1}_{t_{k}\leq t <t_{k+1}}, \qquad k = 0, \ldots,L-1
\end{eqnarray}
and denote 
\begin{eqnarray*}
M^{(L)}_k(h) &=& \left(\mu\left(t_k,h\right) -\frac{1}{2}\sigma^{\prime \prime}(0)Q(t_k,h)\right)\Delta_L+ \Delta V_k^A  \, h + \Delta V_k^b\\
&=&\left(U^{A}_{t_k} \, h+\,U^{b}_{t_k}+ \overline{A}_{t_k}  h + \overline{b}_{t_k}\right)\Delta_L+ \Delta V_k^A  \, h + \Delta V_k^b\\
&=:&\widetilde{\mu}\left(t_k,h\right)\Delta_L+ \Delta V_k^A  \, h + \Delta V_k^b,
\end{eqnarray*}
From~\eqref{eq:euler_maruyama3} we thus have
$$\Delta h_k^{(L)} := h_{k+1}^{(L)}-h_k^{(L)} = \sigma \left(  M_k^{(L)}(h_k^{(L)})\right).$$
For any vector $x\in \mathbb{R}^d$, denote $(x)_i$ as the $i$-th component of $x$ ($i=1,2,\ldots,d$). Further denote $\Delta h_k^{(L),i}$ and $M_{k}^{(L),i}$ the $i$-th element of $\Delta h_k^{(L)}$ and $M_{k}^{(L)}$, respectively. 
Applying a third-order Taylor expansion of $\sigma$ around $0$ with the help of Assumption~\ref{ass:activation}, for $i=1,2,\ldots,d$, we get
\begin{align*}
\Delta h_k^{(L),i} &= \sigma \left(  M^{(L),i}_k(h_k^{(L)})\right) = M^{(L),i}_k(h_k^{(L)}) +\frac{1}{2}\sigma^{\prime \prime}(0)\left(  M^{(L),i}_k(h_k^{(L)})\right)^2+ \frac{1}{6} \sigma^{\prime \prime \prime} (\nu_i)\left(  M^{(L),i}_k(h_k^{(L)}) \right)^3\\
&= \underbrace{{\mu}_i\left(t_{k},h_k^{(L)}\right)\Delta_L + (\Delta V_k^A  \, h_k^{(L)} )_i+ (\Delta V_k^b)_i}_{f_i^{(L)}(k, h_k^{(L)})} + \underbrace{\frac{1}{2}\sigma^{\prime \prime} (0) \left( \left(M_k^{(L), i}(h_k^{(L)})\right)^2 - Q_i(t_k, h_k^{(L)}) \right)}_{N_k^{(L),i}(h_k^{(L)})} + \frac{1}{6} \sigma^{\prime \prime \prime} (\nu_i)\left(  M^{(L),i}_k(h_k^{(L)}) \right)^3 \\
&= f^{(L)}_i(k,h_k^{(L)})+N_k^{(L),i}(h_k^{(L)}) + \frac{1}{6}\sigma^{\prime \prime \prime} (\nu_i)\left(  M^{(L),i}_k(h_k^{(L)}) \right)^3 ,
\end{align*}
with $\abs{\nu_i} < \abs{ M^{(L),i}_k(h_k^{(L)}) }$. The increment of the hidden state $\Delta h_k^{(L),i}$ has two parts: the increment of the Euler-Maruyama scheme $ f^{(L)}_i(k,h_k^{(L)})$ and the residual  $D_k^{(L),i}(h_k^{(L)}) := \frac{1}{6}\sigma^{\prime \prime \prime} (\nu_i)\left(  M^{(L),i}_k(h_k^{(L)}) \right)^3 +N_k^{(L),i}(h_k^{(L)})$. It is clear from here that the Euler-Maruyama scheme of the limiting SDE is different from the ResNet dynamics. Hence classical results on the convergence of discrete SDE schemes cannot be applied directly.

In our analysis it will be more natural to work with the following {\it continuous-time approximation}:
\begin{eqnarray}\label{eq:cta}
\htilde^{(L)}_t :=h_0^{(L)} +\int_0^t \mu\left({ {t_{k_s}}}, \hbarr^{(L)}_s \right) \dd s +\int_0^t \left(\dd V^A_s \, \hbarr^{(L)}_s + \dd V^b_s \right) +\sum_{k\leq Lt} D_k^{(L)}\left( h_k^{(L)} \right),
\end{eqnarray}
where $D_k^{(L)}(h) = \left(D_k^{(L),1}(h),\ldots,D_k^{(L),d}(h) \right)^{\top}$ and $k_s$ is the integer for which $s\in[t_{k_s},t_{k_s+1})$ for a given $s\in[0,1)$. From the above definitions we have $\htilde^{(L)}_{t_k}=\hbarr^{L}_{t_k} = h^{(L)}_k$. 
 
 

 \begin{lemma}[Local Lipschitz condition and uniform integrability]\label{lemma:main} Under the assumptions of Theorem~\ref{thm:H2},
 \begin{enumerate}
     \item  For each $R>0$, there exists a constant $C_R$, depending only on $R$, such that almost surely we have
 \begin{eqnarray}\label{ass:local_lipschitz}
 \norm{\mu(t,x)-\mu(t,y)}^2  \leq C_R \norm{x-y}^2 , \quad \forall x,y \in \mathbb{R}^d \textit{ with } \norm{x} \lor \norm{y} \leq R, \textit{ and } \forall t\in[0,1],
 \end{eqnarray}
 where $\mu$  is defined in~\eqref{eq:case1_mean}.
 \item 
 There exist  constants $p>2$ and $C>0$ such that
 \begin{eqnarray}\label{ass:boundedness.v2}
 \mathbb{E}\left[\sup_{0 \leq t \leq 1} \norm{\htilde^{(L)}_t}^p \right] \lor \mathbb{E}\left[\sup_{0 \leq t \leq 1} \norm{H_t}^p \right] \leq C.
 \end{eqnarray}
 \end{enumerate}
 \end{lemma}
 \begin{remark}\label{rmk:uniform_integrability}
 Note that ~\cite{higham2002strong} assumes the uniform integrability condition for $\htilde_t^{(L)}$, which is difficult to verify in practice.  Here we relax this condition by only assuming the uniform integrability condition for the ResNet dynamics $\{h_k^{(L)} : k=0,\ldots, L \}$, see Assumption~\ref{ass:strong}.  We can then prove~\eqref{ass:boundedness.v2} under Assumption~\ref{ass:strong} and some properties of the It\^o processes.
 \end{remark}
 Lemma~\ref{lemma:main} is proved by first showing $Q(t,x)$ is locally Lipschiz and then by applying Minkowski inequality to $\norm{\hbarr^{(L)}_s-\htilde^{(L)}_s}^{p}$ with $p=\frac{1}{2}p_1>2$, where $p_1$ is defined in \eqref{ass:boundedness}. \\


We are now ready to prove Theorem~\ref{thm:H2}.

\begin{proof}
Let us define two stopping times to utilize the local Lipschitz property of $\mu$:
 \begin{eqnarray}
 \tau_R := \inf \left\{t \geq 0: \norm{\htilde^{(L)}_t} \geq R\right\}, \quad \rho_R := \inf \left\{t \ge 0: \norm{H_t} \geq R\right\},\quad \theta_R := \tau_R \land \rho_R,
 \end{eqnarray}
 and define the approximation errors
 \begin{eqnarray}
 e_1(t):= \htilde^{(L)}_t-H_t, \,\,\text{ and }\,\,
 e_2(t):= \htilde^{(L)}_t-\hbarr^{(L)}_t.
 \end{eqnarray}
 The proof contains two steps. The first step is to show $\lim_{L\rightarrow\infty}\mathbb{E}\left[ \sup_{0 \leq t \leq 1} \norm{e_1(t)}^2\right]=0$ and the second step is to show $\lim_{L\rightarrow\infty}\mathbb{E}\left[ \sup_{0 \leq t \leq 1} \norm{e_2(t)}^2\right]=0$. 

\paragraph{Step 1: $\widetilde{H}$ and $H$ are uniformly close to each other.} Following the idea in \cite{higham2002strong}, we first show that for any $\delta>0$ (to be determined later), by Young's inequality,
 \begin{eqnarray}\label{eq:bound1}
\mathbb{E}\left[ \sup_{0 \leq t \leq 1} \norm{e_1(t)}^2\right] \leq \mathbb{E}\left[ \sup_{0 \leq t \leq 1} \norm{\htilde^{(L)}_{t\wedge \theta_R}-H_{t\wedge \theta_R}}^2\right] +\frac{2^{p+1}\delta C}{p} + \frac{(p-2)2C}{p\delta^{2/(p-2)}R^p},
 \end{eqnarray}
where $C$ and $p$ are defined in~\eqref{ass:boundedness.v2}. Now, we bound the first term on the right-hand side of \eqref{eq:bound1}. Using the definition of the targeted SDE limit in~\eqref{eq:limit_sde}, the continuous-time approximation~\eqref{eq:cta}, and the Cauchy-Schwarz inequality, we get
 \begin{eqnarray*}
\norm{\htilde^{(L)}_{\tR}-H_{\tR}}^2 &\leq & 4 \left[ \int_0^{\tR}\norm{ \mu\left(s,\hbarr^{(L)}_s\right) \dd s-\mu\left(s,H_s\right)}^2 \dd s  \right] + 4 \left [ \int_0^{\tR} \norm{\mu\left({ {t}_{k_s}},\hbarr^{(L)}_s\right) \dd s-\mu\left(s,\hbarr^{(L)}_s\right)}^2 \dd s  \right]  \nonumber\\
&& \,\,+ 4 \, \norm{ \int_0^{\tR}\dd W_s^A \, \left(\hbarr^{(L)}_s-H_s\right) }^2 + \, 4 \, \norm{ \sum_{k\leq L(\tR)} D_k^{(L)}\left( {h}^{(L)}_k \right) }^2.
 \end{eqnarray*}
 Therefore, from the local Lipschitz condition~\eqref{ass:local_lipschitz} and Doob's martingale inequality~\citep{revuz2013continuous}, we have for any $\tau \leq 1$,
 \begin{eqnarray}
 &&\mathbb{E}\left[\sup_{0 \leq t \leq \tau} \norm{\htilde^{(L)}_{\tR}-H_{\tR}}^2 \right] \nonumber \\
 &\leq& 32 \left(C_R+C^2_1\right) \int_0^{\tau}  \mathbb{E} \left[ \sup_{0 \leq r \leq s}\norm{\htilde^{(L)}_{r\wedge\theta_R}-H_{r\wedge\theta_R}}^2\right] \dd s+32\left(C_R+C^2_1\right)\underbrace{\mathbb{E}\int_0^{\tauR} \norm{\hbarr^{(L)}_s-\htilde^{(L)}_s}^2 \dd s}_{\textcircled{1}} \nonumber\\
 && +\, 4  \,\, \underbrace{\mathbb{E}\left[ \int_0^{\tR}\norm{ \mu\left({ {t}_{k_s}},\hbarr^{(L)}_s\right) - \mu\left(s,\hbarr^{(L)}_s\right)}^2 \dd s \right]}_{\textcircled{2}}+ \, 4\underbrace{\mathbb{E}\left[\sup_{0\leq t \leq \tau} \norm{\sum_{k\leq L(\tR)} D_k^{(L)}\left({h}^{(L)}_k\right)}^2\right]}_{\textcircled{3}}. \label{eq:key_bound}
 \end{eqnarray}
 

{\bf Upper bound for~$\textcircled{2}$.}  By the Cauchy–Schwarz inequality, the following holds for almost all $h\in \mathbb{R}^d$:
\begin{eqnarray}
    \norm{\mu(t,h)-\mu(s,h)}^2  \leq C_M \abs{t-s}^{\kappa}\left( 1+ \norm{h}^2 + \norm{h}^4 \right).
 \end{eqnarray}
Under Assumption~\ref{ass:strong}, there exists a constant $\widetilde{C}_0>0$ such that
\begin{eqnarray} \label{eq:bound_hbar}
\mathbb{E}\left[\sup_{0 \leq t \leq 1}\left(\norm{\hbarr^{(L)}_t}^{4}+\norm{\hbarr^{(L)}_t}^{2} \right)\right]  \leq \widetilde{C}_0.
\end{eqnarray} 
Hence by Tonelli's theorem,
\begin{eqnarray}
&&\mathbb{E}\left [ \int_0^{\tR}\norm{\mu\left({ {t}_{k_s}},\hbarr^{(L)}_s \right) - \mu\left(s,\hbarr^{(L)}_s\right)}^2 \dd s\right]\leq \int_0^{1}\mathbb{E}\left[ \norm{\mu\left({ {t}_{k_s}},\hbarr^{(L)}_s\right) - \mu\left(s,\hbarr^{(L)}_s\right)}^2\right]\dd s\nonumber \\
&\leq& (\widetilde{C}_0+1)C_M L \left(\int_0^{1/L} r^{\kappa} \dd r\right) = \frac{(\widetilde{C}_0+1)C_M }{1+\kappa}L^{-\kappa}. \label{eq:tonelli}
\end{eqnarray}

\paragraph{Upper bound for $\textcircled{3}$.}
Define the following discrete filtration $\GG_{k} := \sigma\Big(U_s^A, U_s^A,q_s^A,q_s^b, B_s^{A}, B_s^{b} \,:\,\, s \leq t_{k+1} \Big)$. Note that $h_k^{(L)}$ is $\GG_{k-1}$-measurable but not $\GG_{k}$-measurable. Define for $k=0, \ldots, L-1$ and for $i=1, \ldots, d$:
\begin{eqnarray*}
 X_k^i   &:=& \left( \left(\Delta V_k^A  \, h_{k}^{(L)}\right)_i + \left(\Delta V_k^b \right)_i\right)^2-\mathbb{E} \left[\left. \left( \left(\Delta V_k^A  \, h_{k}^{(L)}\right)_i + \left(\Delta V_k^b\right)_i\right)^2\right\vert \GG_{k-1}\right]\label{eq:x_def}\\
 Y_k^i &:=& \mathbb{E} \left[\left. \left( \left(\Delta V_k^A  \, h_{k}^{(L)}\right)_i + \left(\Delta V_k^b\right)_i\right)^2\right\vert \GG_{k-1}\right] - Q_i\left(t_k, h_k^{(L)} \right)\Delta_L\\
 J_k^i &:=& \widetilde{\mu}_i(t,h)^2 (\Delta_L)^2 + 2 \widetilde{\mu}_i(t,h) \Delta_L \left( \left(\Delta V_k^A  \, h\right)_i + \left(\Delta V_k^b\right)_i\right).
\end{eqnarray*}
We can then deduce the following bound on \textcircled{3} by Cauchy-Schwarz.
\begin{eqnarray}
&&\mathbb{E}\left[\sup_{0\leq t \leq \tau} \abs{\sum_{k\leq L(\tR)} D_k^{(L),i}\left({h}^{(L)}_k\right)}^2\right]\nonumber\\
&\leq& \sigma''(0)^2 \,\, \mathbb{E}\left[\sup_{0\leq t \leq \tau} \abs{\sum_{k\leq Lt} X_k^i\one_{\norm{{h}^{(L)}_k}\leq R}}^2\right] +  \sigma''(0)^2 \,\,\mathbb{E}\left[\sup_{0\leq t \leq \tau} \abs{\sum_{k\leq Lt} Y_k^i\one_{\norm{{h}^{(L)}_k}\leq R}}^2\right] \nonumber \\
&&\quad + \, \sigma''(0)^2 L \sum_{k=0}^{L-1} \mathbb{E}\left[\abs{ J_k^i}^2 \one_{\norm{{h}^{(L)}_k}\leq R}\right]+ \frac{(\sigma'''(\nu_i))^2}{9} L \sum_{k=0}^{L-1} \mathbb{E}\left[\abs{ M^{(L),i}_k\left(h_k^{(L)}\right)}^6 \one_{\norm{{h}^{(L)}_k}\leq R}\right] \label{eq:bound2}
\end{eqnarray}

We provide an upper bound for each of the four terms in \eqref{eq:bound2}. For the first term, denote $\Xtilde_k^i = X_k^i\one\left(\big\lVert {h}^{(L)}_k \big\rVert\leq R\right)$ and $S^i_k = \sum_{k'=0}^k \Xtilde_{k'}^i$ so that $\left(S_k^i\right)_{k=-1, 0,\ldots, L-1}$ is a $\left( \GG_k \right)$-martingale.  Hence, by Doob's martingale inequality, we have
  \begin{equation} \label{eq:doobs-xtilde}
    \E\left[ \sup_{0 \leq t \leq \tau } \abs{ \sum_{k\leq Lt} X_k^i\one_{\norm{{h}^{(L)}_k}\leq R} }^2 \right] = \E\left[ \sup_{0 \leq t \leq \tau } \abs{ S_{\floor{Lt}}^i }^2 \right] \leq 4 \, \E \left[ \abs{ S_{\floor{L \tau}}^i }^2 \right].
  \end{equation}
Fix $k = 0, \ldots, L-1$. For every $i=1, \ldots, d$, we compute the following conditional expectation.
\begin{equation} \label{eq:tower-law}
\E \left[ \left(S_k^i \right)^2 \, \Big\vert \,\, \GG_{k-1} \right]  = \E \left[ \left( S_{k-1}^i \right)^2 + 2 \Xtilde_k^i \sum_{k'=0}^{k-1} \Xtilde_{k'}^i + \left(\Xtilde_{k}^i\right)^2 \, \Bigg\vert \,\, \GG_{k-1} \right] = \left( S_{k-1}^i \right)^2  + \E\left[ \left(\Xtilde_k^i\right)^2 \, \Big\vert \,\, \GG_{k-1} \right]
\end{equation}
The cross-term disappear as $\E\left[\left. \Xtilde_{k}^i \, \right\vert \, \GG_{k-1} \right] = \E\left[\left. X_{k}^i \, \right\vert \, \GG_{k-1} \right]\one\left(\big\lVert {h}^{(L)}_k \big\rVert\leq R\right) = 0$ by definition of $X_k^i$. Furthermore, conditionally on $\GG_{k-1}$ and on $\left\{ \big\lVert {h}^{(L)}_k \big\rVert\leq R \right\}$,  observe that $X_k^i$ is the centered square of a normal random variable whose variance is $\OO(L^{-1})$ uniformly in $k$ by \eqref{eq:bound_ito}, so there exist $C_{R,1} > 0$ depending only on $R$ such that  
\begin{equation*}
\sup_{k=0, \ldots, K-1} \E\left[ \left(\Xtilde_k^i\right)^2 \, \Big\vert \,\, \GG_{k-1} \right] \leq C_{R,1} L^{-2}.
\end{equation*}
Hence, 
plugging  back into \eqref{eq:doobs-xtilde}, 
we obtain
\begin{equation}\label{eq:bound14}
\E\left[ \sup_{0 \leq t \leq \tau } \abs{ \sum_{k\leq Lt} X_k^i\one_{\norm{{h}^{(L)}_k}\leq R} }^2 \right]  \leq 4 C_{R,1} L^{-1}.
\end{equation}

For the second term involving $Y_k^i$, we explicitly compute the conditional expectation using the definition of $V$ in~\eqref{eq:case1_mean} and the definition of $Q$ in~\eqref{eq:Q_thm}.

\begin{align*}
    Y_k^i 
    = \int_{t_k}^{t_{k+1}} \left( \E\left[\left. \Sigma^b_{s, ii} - \Sigma^b_{t_k, ii} \, \right\vert \, \GG_{k-1} \right] + \sum_{j,l=1}^d h^{(L)}_{k, j} h^{(L)}_{k, l}  \E \left[\left. \Sigma^A_{s, ijil} - \Sigma^A_{t_k, ijil} \, \right\vert \, \GG_{k-1} \right] \right) \dd s. 
\end{align*}
Now, we compute directly the following bound by Cauchy-Schwarz, Tonelli and \eqref{eq:continuity_ito} in Assumption \ref{ass:ito}-(ii):
\begin{align}
&\mathbb{E}\left[\sup_{0\leq t \leq \tau} \abs{\sum_{k\leq L(\tR)} Y_k^i\one_{\norm{{h}^{(L)}_k}\leq R}}^2\right] \leq  M(1+R^2)^2 \left( L\int_{0}^{1/L} r^{\kappa/2} \dd r \right)^2 \eqqcolon C_{R,2} L^{-\kappa},\label{eq:bound15}
\end{align}
where $C_{R, 2} > 0$ depends only on $R$.
Moving to the third term of \eqref{eq:bound2} involving $J_k^i$, we get directly from the definition of $V^A$ and $V^b$ that there exists $C_{R,3}>0$ depending only on $R$ such that
\begin{eqnarray}\label{eq:bound10}
\sup_{\|h\|\leq R}\mathbb{E}\left[\abs{ J_k^i}^2 \one_{\norm{h_k^{(L)}}\leq R}\right] \leq C_{R,3} L^{-3}.
\end{eqnarray}

Finally, we bound the fourth term of \eqref{eq:bound2} using Cauchy-Schwarz, Assumption \ref{ass:activation} and property \eqref{eq:bound_ito} of the It\^o processes:
\begin{eqnarray}\label{eq:bound9}
\sigma'''(\nu_i)^2 \sup_{\|h\|\leq R}  \mathbb{E}\left[ \left(M^{(L),i}_k\left(h \right)\right)^6 \right] \leq  m^2\,C_{R,4} L^{-3},
\end{eqnarray}
for some constant $C_{R, 4}>0$ depending only on $R$. Combining the results in  \eqref{eq:bound14}, \eqref{eq:bound15}, \eqref{eq:bound10} and \eqref{eq:bound9}, we have
\begin{align} 
\E\left[\sup_{0\leq t \leq \tau} \abs{\sum_{k\leq L(\tR)} D_k^{(L),i}\left({h}^{(L)}_k\right)}^2\right] &\leq 4 \sigma''(0)^2 C_{R,1}L^{-1} +  \sigma''(0)^2 C_{R, 2} L^{-\kappa} + \sigma''(0)^2 C_{R, 3} L^{-1} + \frac{m^2}{9} C_{R,4} L^{-1} \nonumber \\
&\eqqcolon \frac{C_{R, 5}}{4d}L^{-\kappa} + \frac{C_{R, 6}}{4d} L^{-1}.  \label{eq:bound11}
\end{align}

\paragraph{Upper bound for $\textcircled{1}$.} Given $s\in[0,T\wedge \theta_R)$, denote $k_s$ as the integer for which $s\in[t_k,t_{k_s+1})$. Then
\begin{eqnarray}
\hbarr^{(L)}_s-\htilde^{(L)}_s &=& h_{k_s}^{(L)} - \left( h_{k_s}^{(L)}+\int_{t_{k_s}}^s \mu(s,\hbarr^{(L)}_s)\dd s + \int_{t_{k_s}}^s \left(\dd V_s^A\hbarr_s^{(L)} + \dd V_s^b\right)\right)\nonumber\\
&=&-\mu\left(t_{k_s},h_{k_s}^{(L)} \right) (s-t_{k_s})-\left(V^A_s-V^A_{t_{k_s}}\right) \, h_{k_s}^{(L)}-\left(V^b_s-V^b_{t_{k_s}}\right),\label{eq:Htilde_Hbar}
\end{eqnarray}
and by the Mean-value Theorem and the continuity of $\mu$. Hence
\begin{eqnarray}\label{eq:bound3}
\norm{\hbarr^{(L)}_s-\htilde^{(L)}_s}^2 \leq 3 \norm{\mu\left(t_{k_s},h_{k_s}^{(L)} \right)}^2 (\Delta_L)^2 + 3\norm{h_{k_s}^{(L)}}^2 \norm{V^A_s-V^A_{t_{k_s}}}^2 + 3\norm{V^b_s-V^b_{t_{k_s}}}^2.
\end{eqnarray}

Now, from the local Lipschitz condition \eqref{ass:local_lipschitz}, for $\norm{h} \leq R$ we have almost surely
\begin{eqnarray*}
\norm{\mu(s,h)}^2 \leq 2\left( \norm{\mu(s,h)-\mu(s,0)}^2 + \norm{\mu(s,0)}^2 \right) \leq 2\left(C_R\norm{h}^2 + \norm{\mu(s,0)}^2\right).
\end{eqnarray*}
Hence, 
\begin{eqnarray*}
\eqref{eq:bound3} \leq 4 \left(C_R\norm{h_{k_s}^{(L)}}^2 + \norm{\mu(s,0)}^2 + 1 \right) \left(\Delta_L^2 + \norm{V^A_s-V^A_{t_{k_s}}}^2 + \norm{V^b_s-V^b_{t_{k_s}}}^2 \right).
\end{eqnarray*}
Hence, using \eqref{ass:boundedness.v2} and the Lyapunov inequality \citep{platen2010numerical}, we get
\begin{eqnarray}
\mathbb{E}\int_0^{\tauR}\norm{\hbarr^{(L)}_s-\htilde^{(L)}_s}^2 \dd s
\leq  4  \left(C_R\,C_0^{2/p}+1+\int_0^1 \norm{\mu(s,0)}^2 \dd s\right)\left(\Delta_L^2+4C_1\Delta_L\right). \label{eq:bound12}
\end{eqnarray}

\paragraph{Combining everything:} From \eqref{eq:tonelli}, \eqref{eq:bound11} and \eqref{eq:bound12}, we have in \eqref{eq:key_bound} that
\begin{eqnarray*}
&&\mathbb{E}\left[\sup_{0 \leq t \leq \tau} \norm{\htilde^{(L)}_{\tauR}-H_{\tR}}^2 \right] \leq 128(C_R+C^2_1)  \left(C_R\,C_0^{2/p}+1+\int_0^1 \norm{\mu(s,0)}^2 \dd s\right)\left( L^{-2} + 4C_1 L^{-1} \right) \\ 
&&\qquad + \, \frac{(\widetilde{C}_0+1)C_M }{1+\kappa}L^{-\kappa} + \left(C_{R, 5} L^{-\kappa} + C_{R, 6} L^{-1} \right) +  32 (C_R+C_1^2) \int_0^{\tau} \mathbb{E} \left[ \sup_{0 \leq r \leq s}\norm{\htilde^{(L)}_{r\wedge\theta_R}-H_{r\wedge\theta_R}}^2\right]\dd s.
\end{eqnarray*}
Applying the Grönwall inequality,
\begin{eqnarray}\label{grownwall}
\mathbb{E}\left[\sup_{0 \leq t \leq \tau} \norm{\htilde^{(L)}_{\tauR}-H_{\tR}}^2 \right] \leq C_9 { L^{- \min\{1,\kappa\}}} \Big( C_R^2+C_{R,5}+C_{R,6}+1 \Big) e^{ 32 (C_R+C_1^2)},
\end{eqnarray}
where $C_9$ is a universal constant independent of $L$, $R$ and $\delta$. Combining \eqref{grownwall} with \eqref{eq:bound1}, we have
\begin{eqnarray}
\mathbb{E}\left[ \sup_{0 \leq t \leq 1} \norm{e_1(t)}^2 \right] \leq C_9 { L^{- \min\{1,\kappa\}}} \Big( C_R^2+C_{R,5}+C_{R,6}+1 \Big) e^{ 32 (C_R+C_1^2)} + \frac{2^{p+1}\delta C}{p} + \frac{(p-2)2C}{p\delta^{2/(p-2)}R^p}.\label{eq:bound_final}
\end{eqnarray}
Given any $\epsilon>0$, we can choose $\delta>0$ so that $\frac{2^{p+1}\delta C}{p}<\frac{\epsilon}{3}$, then choose $R$ so that $ \frac{(p-2)2C}{p\delta^{2/(p-2)}R^p}<\frac{\epsilon}{3}$, and finally choose $L$ sufficiently large so that 
\begin{equation*}
C_9{ L^{- \min\{1,\kappa\}}} \Big( C_R^2+C_{R,5}+C_{R,6}+1 \Big)e^{32 (C_R+C_1^2)} \leq \frac{\epsilon}{3}.
\end{equation*}
Therefore in \eqref{eq:bound_final}, we have, 
\begin{eqnarray}\label{eq:bound_e}
\mathbb{E}\left[ \sup_{0 \leq t \leq 1}\norm{e_1(t)}^2\right]  \leq \epsilon.
\end{eqnarray}

\paragraph{Step 2: $\hbarr$ and $\widetilde{H}$ are uniformly close to each other.}
Recall the relationship between $\htilde$ and $\hbarr$ defined in~\eqref{eq:Htilde_Hbar}: by~\eqref{eq:bound_ito} we have almost surely that
\begin{eqnarray*}
\norm{\hbarr^{(L)}_s-\htilde^{(L)}_s}^{2} 
\leq  C_{10}\left( \norm{h_{k_s}^{(L)}}^4+\norm{h_{k_s}^{(L)}}^2+1 \right)\,(\Delta_L)^2 + 3 \left( \norm{h_{k_s}^{(L)}}^{2}  \, \norm{V^A_s-V^A_{t_{k_s}}}^{2} + \norm{V^b_s-V^b_{t_{k_s}}}^{2}\right).
\end{eqnarray*}
Therefore
\begin{eqnarray}
&&\mathbb{E}\left[\sup_{0\leq s  \leq 1}\norm{\hbarr^{(L)}_s-\htilde^{(L)}_s}^{2}\right] 
\leq  C_{10}\left( \mathbb{E}\left[\sup_{0\leq s \leq 1}\norm{h_{k_s}^{(L)}}^4\right]+\mathbb{E}\left[\sup_{0\leq s \leq 1}\norm{h_{k_s}^{(L)}}^2\right]+1 \right)\,(\Delta_L)^2\nonumber\\
&& \qquad+ \, 3\, \left( \left(\mathbb{E}\left[\sup_{0\leq s \leq 1}\norm{h_{k_s}^{(L)}}^{4}\right]  \, \mathbb{E}\left[\sup_{0\leq s \leq 1}\norm{V^A_s-V^A_{t_{k_s}}}^{4}\right]\right)^{1/2} +\mathbb{E}\left[\sup_{0\leq s \leq 1} \norm{V^b_s-V^b_{t_{k_s}}}^{2}\right] \right)\label{eq:sup_2h}.
\end{eqnarray}

By the Power Mean inequality and Doob's martingale inequality,
\begin{eqnarray}\label{eq:bound16}
\mathbb{E}\left[\sup_{0\leq s \leq  1}\norm{V_s^A-V_{t_{k_s}}^A}^4\right] \leq \mathbb{E}\left[\sum_{k=0}^{L-1}\left(\sup_{t_{k}\leq s < t_{k+1}}\norm{V_s^A-V_{t_{k_s}}^A}^4\right)\right] \leq C_{11} \Delta_L.
\end{eqnarray}
Using H\"older's inequality yields
\begin{eqnarray}\label{eq:bound17}
\mathbb{E}\left[\sup_{0\leq s \leq  1}\norm{V_s^A-V_{t_{k_s}}^A}^2\right] \leq \left(\mathbb{E}\left[\sup_{0\leq s \leq  1}\norm{V_s^A-V_{t_{k_s}}^A}^4\right]\right)^{1/2} \leq \sqrt{C_{11}}\Delta^{1/2}_L.
\end{eqnarray}

Combining \eqref{eq:bound_hbar}, \eqref{eq:bound16}, and \eqref{eq:bound17} in \eqref{eq:sup_2h}, we obtain
\begin{eqnarray*}
\mathbb{E}\left[\sup_{0\leq   t \leq 1}\norm{e_2(t)}^{2}\right] = \mathbb{E}\left[\sup_{0\leq t  \leq 1}\norm{\hbarr^{(L)}_t-\htilde^{(L)}_t}^{2}\right] \leq C_{12}\Delta_L^{1/2},
\end{eqnarray*}
for some constant $C_{12}>0$. By choosing $L>(C_{12}/\epsilon)^2$,  we have
\begin{eqnarray}\label{eq:bound_h}
\mathbb{E}\left[\sup_{0\leq t  \leq 1}\norm{e_2(t)}^{2}\right] \leq {\epsilon}.
\end{eqnarray}
Finally, combining \eqref{eq:bound_e} and \eqref{eq:bound_h} leads to the desired result.

\end{proof}